\documentclass[manuscript,screen]{acmart} 

\usepackage{forest}
\usetikzlibrary{shadows}
\usepackage{amsmath}

\usepackage{amssymb}
\usepackage{cleveref}

\usepackage{booktabs}
\usepackage{siunitx}
\usepackage{etoolbox}
\usepackage{tcolorbox}
\usepackage{nicematrix}
\usepackage{multirow}
\usepackage{makecell}
\usepackage{enumitem}
\usepackage{xcolor,pifont}
\usepackage{fontawesome5}
\usepackage{colortbl}
\usepackage{subfig,graphicx}
\usepackage{bbding}
\usepackage{hyperref}

\definecolor{lightcoral}{rgb}{0.94, 0.5, 0.5}
\definecolor{lightgreen}{rgb}{0.56, 0.93, 0.56}
\definecolor{harvestgold}{rgb}{0.85, 0.57, 0.0}
\definecolor{brightlavender}{rgb}{0.75, 0.58, 0.89}
\definecolor{capri}{rgb}{0.0, 0.75, 1.0}
\definecolor{carminepink}{rgb}{0.92, 0.3, 0.26}
\definecolor{celadon}{rgb}{0.67, 0.88, 0.69}
\definecolor{darkpastelgreen}{rgb}{0.01, 0.75, 0.24}

\crefformat{section}{\S#2#1#3} 
\crefformat{subsection}{\S#2#1#3}
\crefformat{subsubsection}{\S#2#1#3}

\newrobustcmd{\B}{\bfseries}
\newcommand*\colourcheck[1]{%
  \expandafter\newcommand\csname #1check\endcsname{\textcolor{#1}{\ding{52}}}%
}
\newcommand*\colourcross[1]{%
  \expandafter\newcommand\csname #1cross\endcsname{\textcolor{#1}{\ding{55}}}%
}

\colourcheck{blue}
\colourcheck{green}
\colourcheck{darkpastelgreen}
\colourcross{blue}
\colourcross{applegreen}
\colourcross{red}

\AtBeginDocument{%
  }

\setcopyright{acmlicensed}
\copyrightyear{2024}
\acmYear{2024}
\acmDOI{XXXXXXX.XXXXXXX}

\acmISBN{978-1-4503-XXXX-X/18/06}

\begin{document}


\title{Towards Lifelong Learning of Large Language Models: A Survey}

\author{Junhao Zheng}
\authornote{The first three authors contributed equally to this research.}
\affiliation{%
  \institution{South China University of Technology}
  \city{Guangzhou}
  \state{Guangdong}
  \country{China}}
\email{junhaozheng47@outlook.com}

\author{Shengjie Qiu}
\authornotemark[1]
\affiliation{%
  \institution{South China University of Technology}
  \city{Guangzhou}
  \state{Guangdong}
  \country{China}}
\email{shengjieqiu6@gmail.com}

\author{Chengming Shi}
\authornotemark[1]
\affiliation{%
  \institution{South China University of Technology}
  \city{Guangzhou}
  \state{Guangdong}
  \country{China}}
\email{secmshi@mail.scut.edu.cn}

\author{Qianli Ma}
\authornote{Corresponding author}
\affiliation{%
  \institution{South China University of Technology}
  \city{Guangzhou}
  \state{Guangdong}
  \country{China}}
\email{qianlima@scut.edu.cn}

\renewcommand{\shortauthors}{Junhao Zheng, Shengjie Qiu, Chengming Shi, and Qianli Ma}

\begin{abstract}

As the applications of large language models (LLMs) expand across diverse fields, their ability to adapt to ongoing changes in data, tasks, and user preferences becomes crucial. Traditional training methods with static datasets are inadequate for coping with the dynamic nature of real-world information. Lifelong learning, or continual learning, addresses this by enabling LLMs to learn continuously and adapt over their operational lifetime, integrating new knowledge while retaining previously learned information and preventing catastrophic forgetting.
Our survey explores the landscape of lifelong learning, categorizing strategies into two groups based on how new knowledge is integrated: Internal Knowledge, where LLMs absorb new knowledge into their parameters through full or partial training, and External Knowledge, which incorporates new knowledge as external resources like Wikipedia or APIs without updating model parameters. The key contributions of our survey include: (1) Introducing a novel taxonomy to categorize the extensive literature of lifelong learning into 12 scenarios; (2) Identifying common techniques across all lifelong learning scenarios and classifying existing literature into various technique groups; (3) Highlighting emerging techniques such as model expansion and data selection, which were less explored in the pre-LLM era.
Resources are available at \url{https://github.com/qianlima-lab/awesome-lifelong-learning-methods-for-llm}.

\end{abstract}



\received{10 June 2024}

\maketitle

\hypersetup{
colorlinks=false,
linkcolor=black,
filecolor=black,      
urlcolor=black,
citecolor=black,
}

\section{Introduction}

\begin{figure*}[!t]
\centering
\includegraphics[width=0.7\linewidth]{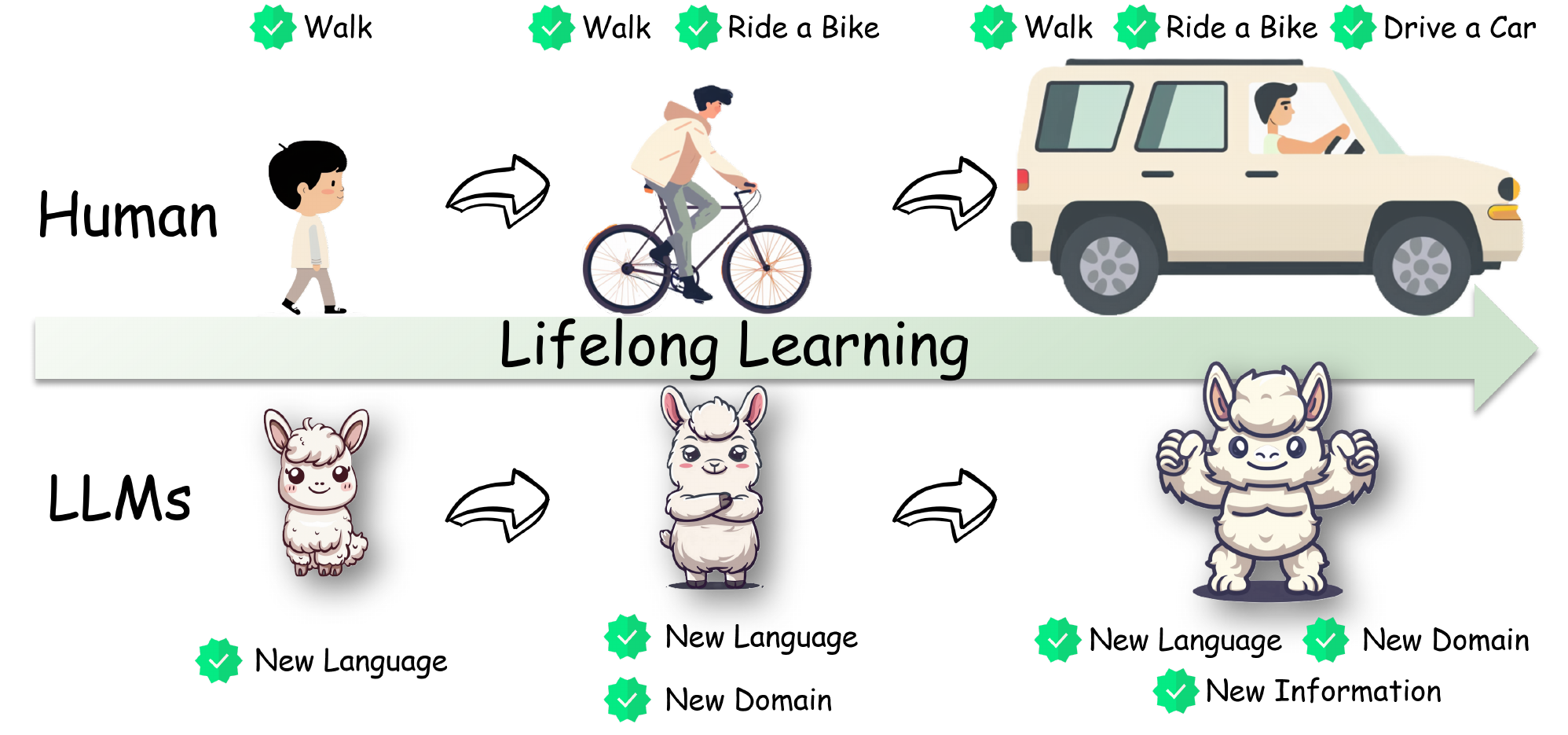}
\caption{
An illustration of lifelong learning: humans can incrementally learn new skills such as walking, riding a bike, and driving a car. Similarly, lifelong learning aims to equip LLMs with new languages, domain knowledge, and information.
}
\label{fig:illustration_lifelong_learning}
\end{figure*}

As the applications of large language models (LLMs) \cite{achiam2023gpt,reid2024gemini,chowdhery2023palm,touvron2023llama,zeng2022glm} expand across diverse fields, the ability of these models to adapt to ongoing changes in data, tasks, and user preferences becomes crucial. Traditional training methods, which rely on static datasets to train LLMs, are increasingly inadequate for coping with the dynamic nature of real-world information \cite{zhang2023large}. Lifelong learning \cite{wang2024comprehensive} (a.k.a., continual learning, incremental learning), or the capability of LLMs to learn continuously and adaptively over their operational lifetime \cite{shi2024continual}, addresses this challenge by integrating new knowledge while retaining previously learned information, thereby preventing the problem of catastrophic forgetting \cite{mccloskey1989catastrophic}.
An illustration of lifelong learning is provided in Figure \ref{fig:illustration_lifelong_learning}.

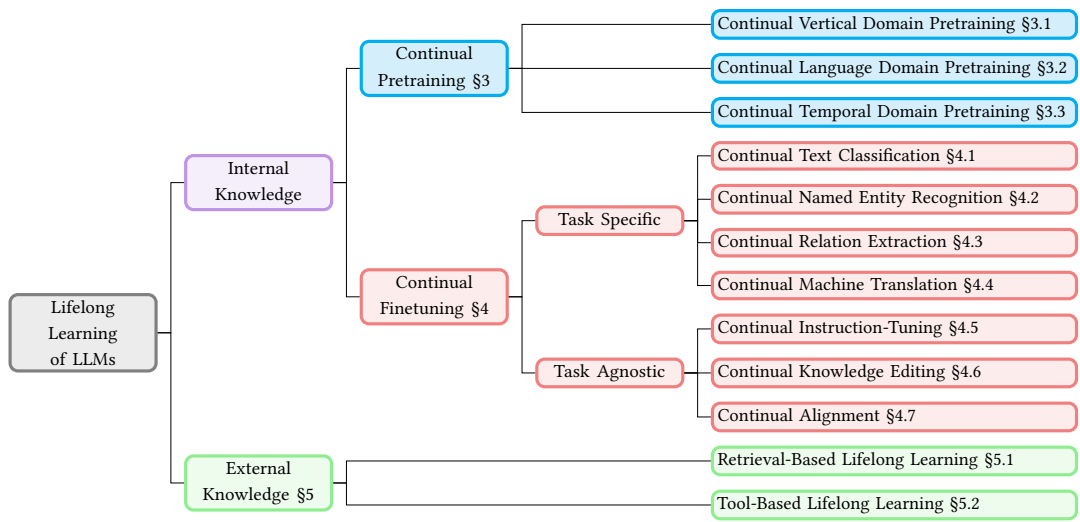
\begin{figure*}[!t]
\centering
\tikzset{
        my node/.style={
            draw,
            align=center,
            thin,
            text width=1.8cm, 
            rounded corners=3,
        },
        my leaf/.style={
            draw,
            align=left,
            thin,
            text width=4.5cm, 
            rounded corners=3,
        }
}
\forestset{
  every leaf node/.style={
    if n children=0{#1}{}
  },
  every tree node/.style={
    if n children=0{minimum width=1em}{#1}
  },
}
\begin{forest}
    for tree={%
        every leaf node={my leaf, font=\footnotesize},
        every tree node={my node, font=\small, l sep-=4.5pt, l-=1.pt},
        anchor=west,
        inner sep=2pt,
        l sep=10pt, 
        s sep=5pt, 
        fit=tight,
        grow'=east,
        edge={ultra thin},
        parent anchor=east,
        child anchor=west,
        if n children=0{tier=last}{},
        edge path={
            \noexpand\path [draw, \forestoption{edge}] (!u.parent anchor) -- +(5pt,0) |- (.child anchor)\forestoption{edge label};
        },
        if={isodd(n_children())}{
            for children={
                if={equal(n,(n_children("!u")+1)/2)}{calign with current}{}
            }
        }{}
    }
    [\footnotesize Lifelong Learning of LLMs, draw=gray, color=gray!100, fill=gray!15, very thick, text=black 
        [\footnotesize Internal Knowledge, color=brightlavender!100, fill=brightlavender!15, very thick, text=black
            [\footnotesize Continual Pretraining \cref{sec:continual_pretraining}, color=cyan!100, fill=cyan!15, very thick, text=black
                [Continual Vertical Domain Pretraining \cref{sec:continual_vertical_domain_pretraining}, color=cyan!100, fill=cyan!15, very thick, text=black, tier=Task, text width=4.7cm
                ]
                [Continual Language Domain Pretraining \cref{sec:continual_language_domain_pretraining}, color=cyan!100, fill=cyan!15, very thick, text=black, tier=Task, text width=4.7cm
                ]
                [Continual Temporal Domain Pretraining \cref{sec:continual_temporal_domain_pretraining}, color=cyan!100, fill=cyan!15, very thick, text=black, tier=Task, text width=4.7cm
                ]
            ]
            [\footnotesize Continual Finetuning \cref{sec:continual_finetuning}, color=lightcoral!100, fill=lightcoral!15, very thick, text=black
                [\footnotesize Task Specific, color=lightcoral!100, fill=lightcoral!15, very thick, text=black
                    [Continual Text Classification \cref{sec:continual_text_classification}, color=lightcoral!100, fill=lightcoral!15, very thick, text=black, tier=Task, text width=4.7cm
                    ]
                    [Continual Named Entity Recognition \cref{sec:continual_named_entity_recognition}, color=lightcoral!100, fill=lightcoral!15, very thick, text=black, tier=Task, text width=4.7cm
                    ]
                    [Continual Relation Extraction \cref{sec:continual_relation_extraction}, color=lightcoral!100, fill=lightcoral!15, very thick, text=black, tier=Task, text width=4.7cm
                    ]
                    [Continual Machine Translation \cref{sec:continual_machine_translation}, color=lightcoral!100, fill=lightcoral!15, very thick, text=black, tier=Task, text width=4.7cm
                    ]
                ]
                [\footnotesize Task Agnostic, color=lightcoral!100, fill=lightcoral!15, very thick, text=black
                    [Continual Instruction-Tuning \cref{sec:continual_instruction_tuning}, color=lightcoral!100, fill=lightcoral!15, very thick, text=black, tier=Task, text width=4.7cm
                    ]
                    [Continual Knowledge Editing \cref{sec:continual_knowledge_editing}, color=lightcoral!100, fill=lightcoral!15, very thick, text=black, tier=Task, text width=4.7cm
                    ]
                    [Continual Alignment \cref{sec:continual_alignment}, color=lightcoral!100, fill=lightcoral!15, very thick, text=black, tier=Task, text width=4.7cm
                    ]
                ]
            ]
        ]
        [\footnotesize External Knowledge \cref{sec:external_knowledge}, color=lightgreen!100, fill=lightgreen!15, very thick, text=black 
            [Retrieval-Based Lifelong Learning \cref{sec:retrieval_based_lifelong_learning}, color=lightgreen!100, fill=lightgreen!15, very thick, text=black, tier=Task, text width=4.7cm
            ]
            [Tool-Based Lifelong Learning \cref{sec:tool_based_lifelong_learning}, color=lightgreen!100, fill=lightgreen!15, very thick, text=black, tier=Task, text width=4.7cm
            ]
        ]
    ]
\end{forest}
\caption{
Taxonomy of lifelong learning methods for LLMs.
}
\label{fig:taxonomy_of_methods}
\end{figure*}

This survey delves into the sophisticated landscape of lifelong learning, categorizing strategies into two primary groups based on how new knowledge is integrated: Internal Knowledge and External Knowledge. Each category encompasses distinct approaches that collectively aim to enhance the adaptability and effectiveness of LLMs in various scenarios. We provide the taxonomy of lifelong learning methods for LLMs in Figure~\ref{fig:taxonomy_of_methods}. 

The Internal Knowledge group, where LLMs absorb new knowledge into their parameters through full or partial training, includes strategies such as continual pretraining \cite{Lifelongpretraining,elle,evaluating,continuallearning,timelms} and continual finetuning \cite{huang2021continual,monaikul2021continual,wang2019sentence,sun2019lamol,lin2023mitigating,CKEsurvey,zheng2024concept}. For example, in industry applications, continual vertical domain pretraining \cite{corpusbrain,large} is commonly adopted, where companies frequently retrain their LLMs using domain-specific data from sectors like finance \cite{yang2023fingpt}. Although this enhances performance in specialized areas, it risks diminishing the model’s broader knowledge base, illustrating the challenges of maintaining a balance between specialized adaptation and general knowledge retention. Continual finetuning covers methods tailored to specific scenarios—such as text classification \cite{huang2021continual}, named entity recognition \cite{monaikul2021continual}, relation extraction \cite{wang2019sentence}, and machine translation \cite{cao2021continual}—as well as task-agnostic methods like instruction tuning \cite{sun2019lamol}, alignment \cite{lin2023mitigating}, and knowledge editing \cite{CKEsurvey}. Additionally, reinforcement learning with human feedback \cite{stiennon2020learning} is employed in continual alignment to ensure that LLMs adhere to human values like safety and politeness \cite{qi2023fine,lermen2023lora}, highlighting the trade-off known as the ``alignment tax'' \cite{lin2023mitigating}, where focusing too narrowly on specific values can lead to a degradation of the model’s general capabilities.

External Knowledge, which incorporates new knowledge as external resources like Wikipedia or APIs without updating model parameters, includes retrieval-based \cite{Densepassage} and tool-based lifelong learning \cite{qin2023tool}, which leverage external data sources and computational tools to extend the model's capabilities. Retrieval-based strategies, such as retrieval-augmented generation \cite{Densepassage,Interleavingretrievalwith,Self-rag,Treeofclarifications,Activeretrievalaugmented}, enhance text generation by providing contextually relevant, accurate, and latest information from external databases such as Wikipedia, ensuring the model's outputs remain relevant over time. Meanwhile, tool-based learning draws parallels to human tool use \cite{allen2020rapid}, where models learn to utilize external computational tools, thus broadening their problem-solving capabilities without direct modifications to their core knowledge base.

Through a detailed examination of these groups and their respective categories, this paper aims to highlight the integration of lifelong learning capabilities into LLMs, thereby enhancing their adaptability, reliability, and overall performance in real-world applications. By addressing the challenges associated with lifelong learning and exploring the innovations in this field, this survey seeks to contribute to the ongoing development of more robust and versatile LLMs capable of thriving in an ever-evolving digital landscape.

\textbf{Differences between this survey and existing ones.}\quad
Lifelong learning has become an increasingly popular research topic in recent years.
Massive surveys have explored the lifelong learning of neural networks \cite{ke2022continual, biesialska2020continual, zhou2024continual, wang2024comprehensive, zhang2023large, wu2024continual, shi2024continual, yuan2023survey, tian2024continual,febrinanto2023graph,de2021continual,yuan2023continual,parisi2019continual,shaheen2022continual,menezes2023continual,yang2024continual}. 
Most of the existing surveys primarily focus on the lifelong learning of Convolutional Neural Networks (CNNs) \cite{biesialska2020continual,wang2024comprehensive, zhou2024continual, yuan2023survey, de2021continual,parisi2019continual,menezes2023continual}.
They examined various scenarios of lifelong learning of CNNs, including image classification \cite{biesialska2020continual,wang2024comprehensive, zhou2024continual, de2021continual,parisi2019continual}, segmentation \cite{yuan2023survey}, objection detection \cite{menezes2023continual}, autonomous systems \cite{shaheen2022continual}, robotics \cite{lesort2020continual}, and the smart city \cite{yang2024continual}.
Besides, several surveys explored the lifelong learning of Graph Neural Network \cite{tian2024continual,zhang2024continual,febrinanto2023graph,yuan2023continual}.
However, only a small amount of literature focuses on lifelong learning of language models \cite{ke2022continual, biesialska2020continual,wu2024continual, shi2024continual,jovanovic2024trends,zhang2023large}.
Biesialska et al. \cite{biesialska2020continual} is an early survey about lifelong learning in Natural Language Processing (NLP). However, they only focus on lifelong learning of word and sentence representations, language modeling, question and answering, text classification, and machine translation.
Ke et al. \cite{ke2022continual} focus on lifelong learning scenarios, including sentiment classification, named entity recognition, and summarization.
They also discuss the techniques for knowledge transfer and inter-task class separation for lifelong learning.
\cite{wu2024continual, shi2024continual,jovanovic2024trends,zhang2023large} are four recent surveys closely related to this research.
Zhang et al. \cite{zhang2023large} provide a comprehensive review of techniques in aligning LLMs with the ever-changing world knowledge, including continual pretraining, knowledge editing, and retrieval augmented generation.
Wu et al. \cite{wu2024continual} revisit lifelong learning from three aspects, including continual pretraining, continual instruction tuning, and continual alignment.
Shi et al. \cite{shi2024continual} examine the lifelong learning of LLMs from two directions including vertical direction (or vertical continual learning), i.e., a continual adaptation from general to specific capabilities, and horizontal direction (or horizontal continual learning), i.e., continual adaptation across time and domains.
Jovanovic et al. \cite{jovanovic2024trends} review several real-time learning paradigms, including continual learning, meta-learning, parameter-efficient learning, and mixture-of-experts learning.
Although recent surveys \cite{wu2024continual, shi2024continual,jovanovic2024trends,zhang2023large} collects the latest literature for lifelong learning, none of them covers the scenarios including continual text classification, continual named entity recognition, continual relation extraction, and continual machine translation, and have little discussion about continual alignment, continual knowledge editing, tool-based lifelong learning and retrieval-based lifelong learning.
\textbf{To our best knowledge, we are the first survey to provide a thorough and systematic examination of lifelong learning methods for LLMs from 12 scenarios.}

\textbf{Contributions of this survey.}\quad The key contributions of our survey are:
\begin{itemize}
    \item \textbf{Novel Taxonomy}: We introduce a detailed and structured framework for categorizing the extensive literature of lifelong learning into 12 scenarios (shown in Figure~\ref{fig:taxonomy_of_methods}).
    \item \textbf{Common Techniques}: We identify common techniques across all lifelong learning scenarios in Section~\ref{sec:overview_common_techniqes} and classify existing literature into various technique groups within each scenario (e.g., Table~\ref{tab:TC_NER_methods}, \ref{tab:RE_MT_methods}, \ref{tab:task_agnostic_methods}).
    \item \textbf{Future Directions}: We highlight several emerging techniques, such as model expansion (section~\ref{sec:continual_vertical_domain_pretraining_model_expansion}) and data selection (section~\ref{sec:continual_vertical_domain_pretraining_data_selection}), that were less explored in the pre-LLM era.
\end{itemize}

\textbf{Organization of this survey.}\quad
The remainder of this paper is organized as follows.
Section~\ref{sec:overview} introduces the problem formulation, evaluation metrics, common techniques, benchmarks, and datasets for lifelong learning. 
Section~\ref{sec:continual_pretraining}, Section~\ref{sec:continual_finetuning}, and Section~\ref{sec:external_knowledge} examine the existing techniques for continual pretraining, continual finetuning, and external-knowledge-based lifelong learning.
Section~\ref{sec:discussion_conclusion} discusses the existing challenges, current trends, and future directions for lifelong learning with LLMs and concludes this survey. 
\section{Overview of Lifelong Learning} \label{sec:overview} 

\subsection{Problem Formulation}
Formally, lifelong learning aims to learn a language model $f_\theta:\mathbf{x}\rightarrow \mathbf{y}$ from the sequence of tasks $\{\mathcal{D}^{(1)},\mathcal{D}^{(2)},\cdots,\mathcal{D}^{(T)}\}$, where the $t$-th task $\mathcal{D}^{(t)} = \{ (\mathbf{x}^{(t)},\mathbf{y}^{(t)})\}$ contains input $\mathbf{x}^{(t)}$ and target output $\mathbf{y}^{(t)}$.  
The input $\mathbf{x}$ and $\mathbf{y}$ are both natural languages.
For generation tasks such as question and answering, $\mathbf{x}$ and $\mathbf{y}$ represent questions and answers.
In machine translation,  $\mathbf{x}$ and $\mathbf{y}$ represent the source and target language.
In text classification,  $\mathbf{x}$ and $\mathbf{y}$ represent the input text and the class label name.
In pretraining tasks for autoregressive language models, $\mathbf{x}$ represents a sequence of tokens $[x_1, x_2, \cdots, x_{n-1}]$, and $\mathbf{y}$ represents the corresponding sequence where each token is the next token in the original input, $[x_2, x_3, \cdots, x_n]$.

\subsection{Evaluation Metrics}
The assessment of continual learning's effectiveness can be approached from three angles: the overall performance of all tasks learned so far, the stability of previously learned tasks, and the plasticity to new tasks.
\begin{itemize}
    \item \textbf{Overall Measurement}: (1) \emph{average accuracy} ($\mathrm{AA}$, higher is better) is computed as the average performance on all tasks learned so far. Formally, the average accuracy when the model has learned $t$ tasks is defined as follows:
\begin{equation}
    \mathrm{AA}_t = \frac{1}{t}\Sigma_{i=1}^{t}a_{t,i},
\end{equation} where $a_{t,i}$ is the performance score on task $i$ when the model has learned $t$ tasks. We suppose that the performance score is higher when the performance is better.
(2) \emph{average incremental accuracy} ($\mathrm{AIA}$, higher is better) is computed as the average of the average accuracy after learning each task. Suppose there are a total of $T$ tasks, we have
\begin{equation}
    \mathrm{AIA} = \frac{1}{T}\Sigma_{t=1}^{T}AA_{t}.
\end{equation}
Compared to $\mathrm{AA}$, $\mathrm{AIA}$ captures the historical variation when learning each task.
    \item \textbf{Stability Measurement}: (1) \emph{forgetting measure} ($\mathrm{FGT}$, lower is better) evaluates the average performance drop of each old task. The performance drop is defined as the difference between its maximum performance obtained previously and its current performance. Formally, the forgetting measure after learning $t$ tasks is defined as follows:
\begin{equation}
    \mathrm{FGT}_t = \frac{1}{t-1}\Sigma_{i=1}^{t-1}[ max_{j\in\{i,i+1,\cdots,t\}}(\{a_{j,i}\}_j) - a_{t,i}],
\end{equation}
where $max_{j\in\{i,i+1,\cdots,t\}}(\{a_{j,i}\}_j)$ represents the maximum performance of task $i$ after task $i$ has been learned, and $a_{t,i}$ represents the performance of task $i$ after learning $t$ tasks.
(2) \emph{backward transfer} ($\mathrm{BWT}$, higher is better) evaluates the average performance change of each old task. The performance change is defined as the difference between its current performance and its performance at the time the task was initially learned. Formally, the backward transfer after learning $t$ tasks is defined as follows:
\begin{equation}
    \mathrm{BWT}_t = \frac{1}{t-1}\Sigma_{i=1}^{t-1} (a_{t,i} - a_{i,i}).
\end{equation}
    \item \textbf{Plasticity Measurement}: \emph{forward transfer} ($\mathrm{FWD}$, higher is better) evaluates the average enhancement in performance on each newly acquired task. This metric calculates the improvement as the difference between the task's initial performance when first learned and the performance of a model that starts with no prior knowledge and is trained only in this task. Formally, the forward transfer after learning $t$ tasks is defined as follows:
\begin{equation}
    \mathrm{FWD}_t = \frac{1}{t-1}\Sigma_{i=2}^{t} (a_{i,i} - \tilde{a_{i}}),
\end{equation}
where $\tilde{a_{i}}$ is the performance of a randomly-initialized model trained on $\mathcal{D}^{(i)}$ only.
\end{itemize}

\begin{figure}
    \centering

    \subfloat[Replay-Based]{
        \includegraphics[width=0.20\linewidth]{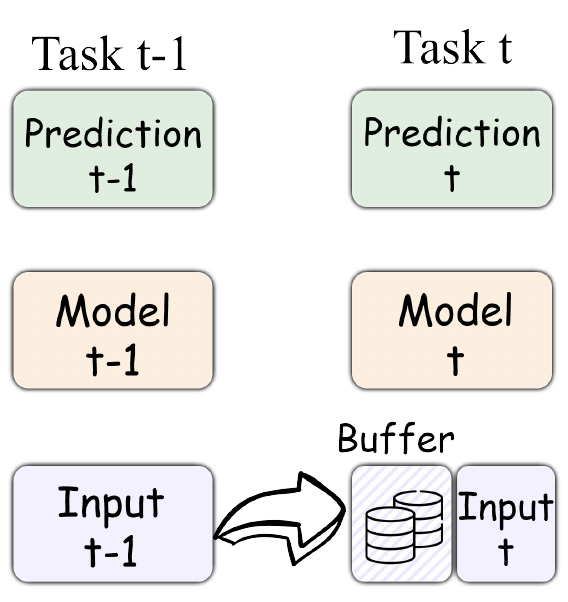}
    }
    \hspace{0.25cm}
    \subfloat[Regularization-Based]{
        \includegraphics[width=0.20\linewidth]{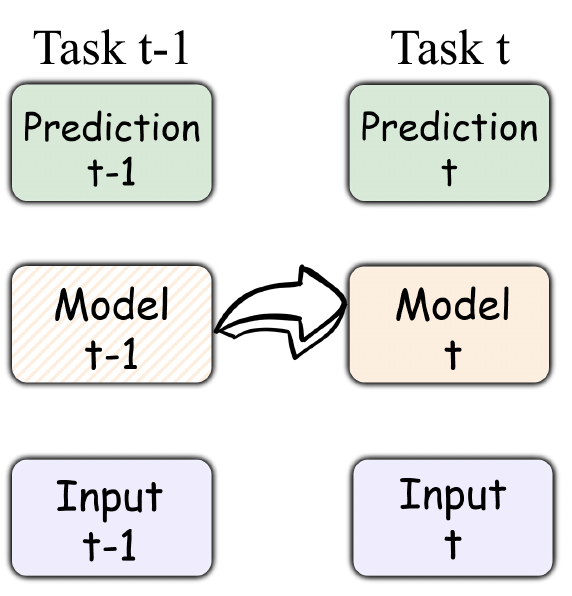}
    }
    \hspace{0.25cm}
    \subfloat[Architecture-Based]{
        \includegraphics[width=0.20\linewidth]{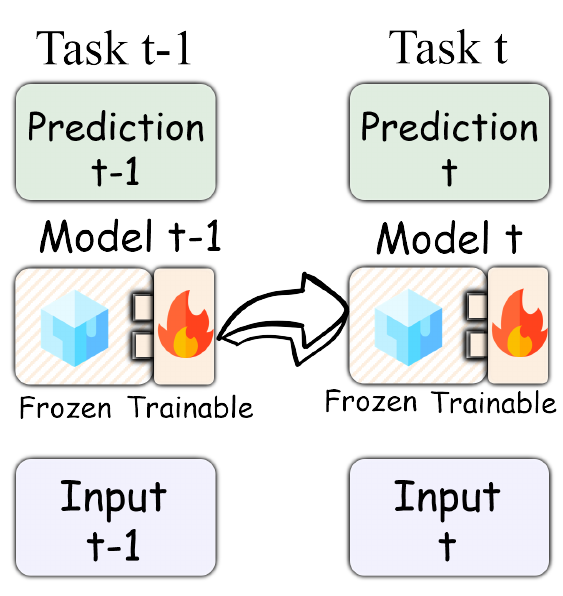}
    }
    \hspace{0.25cm}
    \subfloat[Distillation-Based]{
        \includegraphics[width=0.20\linewidth]{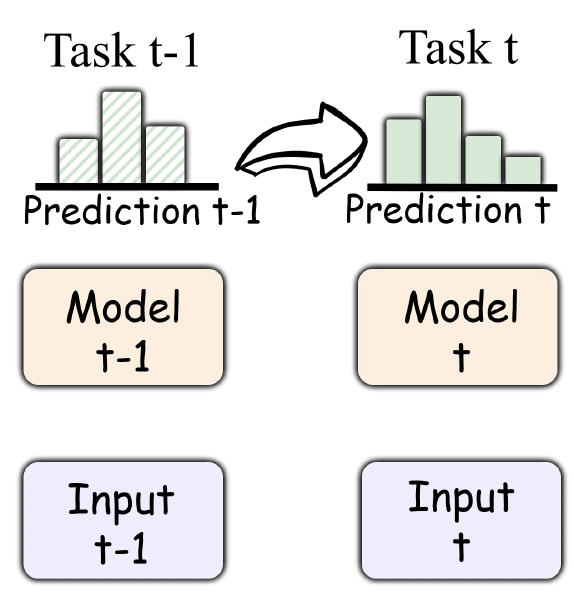}
    }
    
    \caption{Four categories of common techniques for lifelong learning with LLMs.}
    \label{fig:four_categories_methods}
\end{figure}

\subsection{Common Techniques} \label{sec:overview_common_techniqes}
The existing techniques for lifelong learning can be roughly divided into four categories: \emph{replay-based methods}, \emph{regularization-based methods}, \emph{architecture-based methods}, and \emph{distillation-based methods}.
An illustration of four categories of lifelong learning methods is provided in Figure~\ref{fig:four_categories_methods}.

\subsubsection{Replay-based methods}
Replay-based methods are primarily categorized into Experience Replay and Generative Replay, based on how the replay data is obtained.

\begin{itemize}
    \item \textbf{Experience Replay}: This approach involves retaining a subset of previously encountered data or simpler representations of that data, which are periodically reintegrated during the training of new tasks. This technique helps sustain the model's performance on prior tasks by re-exposing it to old data, reinforcing the existing knowledge. For example, in the context of continual pretraining, \cite{domain,elle,Lifelongpretraining,corpusbrain} systematically reintroduces domain-specific datasets during training phases to refresh the model’s memory and stabilize its learning across various domains.
    \item \textbf{Generative Replay}: Instead of storing actual data, this method generates new data samples that emulate old data, using either the model itself or a separate generative model. This approach facilitates continuous learning without the need to retain large volumes of actual data, optimizing memory use and potentially protecting privacy. Within the scope of continual instruction tuning, several innovative methods exemplify generative replay: LAMOL \cite{sun2019lamol}, LFPT5 \cite{qin2021lfpt5}, PCLL \cite{zhao2022prompt} and SSR \cite{huang2024mitigating} generates pseudo instances that are conditioned on natural language cues.
\end{itemize}

\subsubsection{Regularization-Based Methods}

Based on the component they regularize, methods employing regularization can be broadly categorized into weight regularization and feature regularization:

\begin{itemize}
    \item \textbf{Weight Regularization}: This technique penalizes changes to the weights that were important for previous tasks, thus preserving the performance on those tasks. Common strategies include L2 Regularization, which imposes a penalty on the square of the weights to deter large changes; Elastic Weight Consolidation (EWC) \cite{kirkpatrick2017overcoming}, selectively penalizing changes to weights that are critical for past tasks based on their calculated importance; Memory Aware Synapses (MAS) \cite{aljundi2018memory}, which dynamically adjusts the penalty according to the parameter's sensitivity to changes in task performance. Additionally, RecAdam \cite{chen2020recall} incorporates ideas from EWC, introducing a regularization to the pretrained weights with an annealing coefficient to gradually integrate the importance of past knowledge.
    \item \textbf{Feature Regularization}: This method involves constraining the features extracted by the model so that new learning does not significantly interfere with the features learned from previous tasks. Techniques such as IDBR \cite{huang2021continual} and CPFD \cite{zhang2023continual} apply constraints directly on the features to ensure that the activation patterns remain stable across tasks, maintaining a consistent representation space. 
\end{itemize}

\begin{figure}
    \centering

    \subfloat[Prompt Tuning]{
        \includegraphics[width=0.15\linewidth]{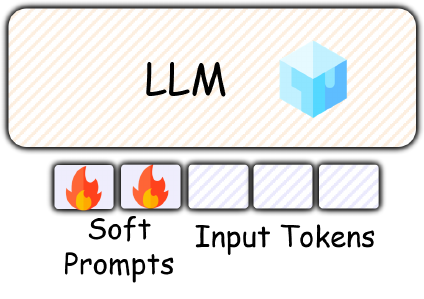}
    }
    \hspace{0.02cm}
    \subfloat[Prefix Tuning]{
        \includegraphics[width=0.15\linewidth]{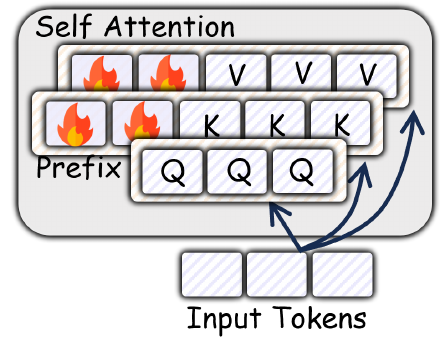}
    }
    \hspace{0.02cm}
    \subfloat[LoRA]{
        \includegraphics[width=0.15\linewidth]{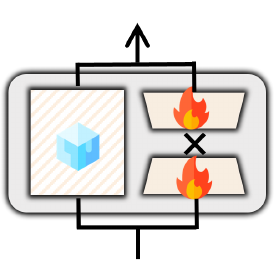}
    }
    \hspace{0.02cm}
    \subfloat[Adapters]{
        \includegraphics[width=0.15\linewidth]{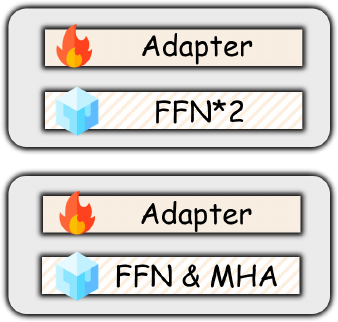}
    }
    \hspace{0.02cm}
    \subfloat[Mixture of Experts]{
        \includegraphics[width=0.15\linewidth]{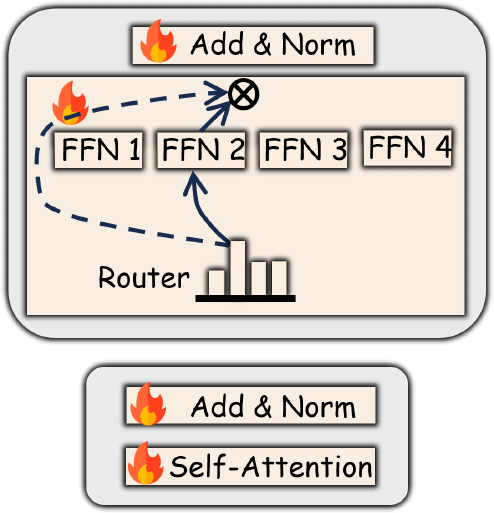}
    }
    \hspace{0.05cm}
    \subfloat[Model Expansion]{
        \includegraphics[width=0.15\linewidth]{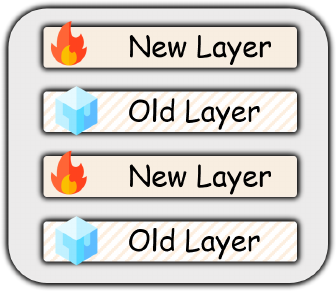}
    }

    \caption{Six categories of architecture-based lifelong methods for LLMs.}
    \label{fig:six_categories_architecture_based_methods}
\end{figure}

\subsubsection{Architecture-Based Methods}

Architecture-based methods in lifelong learning focus on adapting the structure of models to seamlessly integrate new tasks while minimizing disruption to previously acquired knowledge. 
These techniques are particularly vital for existing large language models, such as LLaMA-65B \cite{touvron2023llama}, GLM-130B \cite{zeng2022glm}, PaLM-540B \cite{chowdhery2023palm}, and GPT-4 \cite{achiam2023gpt}, as fully fine-tuning such large-scale models demands extensive computational resources. 
Given these constraints, it is both practical and necessary to pursue efficient and cost-effective lifelong learning strategies. Below is a concise overview of six prominent architecture-based methods for lifelong learning and an illustration is provided in Figure~\ref{fig:six_categories_architecture_based_methods}:

\begin{itemize}
    \item \textbf{Prompt Tuning} \cite{lester2021power}: In Prompt Tuning, trainable task-specific prompts are inserted at the model's input layer to steer its responses towards desired outcomes. This method operates by embedding these prompts directly into the input sequence, affecting only the initial processing of input data. Examples include L2P \cite{wang2022learning}, CODA-Prompt \cite{smith2023coda}, SAPT \cite{modulesapt}, ConvPrompt \cite{roy2024convolutional}, Q-Tuning \cite{guo2024q}, and Fwd-Prompt \cite{zheng2024beyond}.
    \item \textbf{Prefix Tuning} \cite{li2021prefix}: This method involves prepending a set of trainable parameters, known as prefixes, to each layer of the transformer model. These prefixes act as contextual modifications that adjust the model's behavior for specific tasks. Prefix Tuning influences multiple layers of the model, in contrast to Prompt Tuning. Notable implementations include EPI \cite{wang2023rehearsal} and MoCL \cite{wang2024rehearsal}.
    \item \textbf{LoRA} (Low-Rank Adaptation) \cite{hu2021lora}: LoRA integrates low-rank matrices within certain layers of a pre-trained model to adapt its functionality without comprehensive retraining. It allows for targeted adjustments to specific model components. Methods utilizing LoRA include Lee et al. \cite{lee2022plug}, C-LoRA \cite{smith2023continual}, ConPET \cite{song2023conpet}, GLRL \cite{zhao2023inrank}, O-LoRA \cite{wang2023orthogonal}, CoLoR \cite{wistuba2023continual}, InfLoRA \cite{liang2024inflora}, SAPT \cite{modulesapt}, MoRAL \cite{moral}, EKFAC \cite{chen2024bayesian}, and I-LoRA \cite{ren2024analyzing}.
    \item \textbf{Adapters} \cite{houlsby2019parameter}: These are small, two-layer feed-forward neural networks with a bottleneck structure, inserted between the layers of the existing model architecture. They allow the model to acquire new capabilities while preserving the original pre-trained parameters intact. Examples include CPT \cite{ke2022continual}, LAFT-URIEL \cite{parameter}, DMEA \cite{qin2023lifelong}, TSS \cite{ke2023sub}, HOP \cite{michieli2024hop}, and SEMA \cite{wang2024self}.
    \item \textbf{Mixture of Experts} (MoE) \cite{shazeer2016outrageously}: MoE approaches utilize a gating mechanism to dynamically select from a set of expert feed-forward neural networks during inference, based on the task at hand. This allows the model to specialize certain parts of its architecture to specific types of tasks, enhancing performance and scalability. Examples include DEMix \cite{demix} and ModuleFormer \cite{moduleformer}.
    \item \textbf{Model Expansion} \cite{chen2015net2net}: This category includes techniques that either reuse existing model components or expand the model architecture to accommodate new information and tasks. This can involve adding new layers or modules, or scaling existing ones to increase the model's capacity and flexibility. Notable methods include bert2BERT \cite{bert2bert}, Wang et al. \cite{wang2023data}, LLaMA Pro \cite{llama}, and SOLAR \cite{solar}.
\end{itemize}

\subsubsection{Distillation-Based Methods}

Based on the source of the distilled targets, distillation-based methods can be categorized into three groups: new data, old data, and pseudo-old data:
\begin{itemize}
    \item \textbf{Distillation from New Data}: These techniques involve the student model learning directly from new tasks under the guidance of a teacher model with new data. Representative methods include Learning without Forgetting (LwF) \cite{li2017learning}, where the model adapts to new classes without forgetting older ones. In continual named entity recognition, the overlap between new and old entities is addressed by methods like ExtendNER \cite{monaikul2021continual} and CFNER \cite{zheng2022distilling}, which use the old model to generate pseudo soft labels for ``Other'' tokens, aiding the learning of new entities while maintaining old knowledge. Additionally, in continual machine translation, methods such as Cao et al. \cite{cao2021continual}, COKD \cite{shao2022overcoming}, LFR \cite{gu2022continual}, and CKD \cite{zhang2023continual} employ distillation strategies focusing on new data.
    \item \textbf{Distillation from Old Data}: This category uses old data, which is typically stored in memory, to guide the student model through the outputs of a teacher model. Examples include CRN \cite{bai2022incremental}, CRL \cite{CRL}, SCKD \cite{SCKD}, and CEAR \cite{CEAR}.
    \item \textbf{Distillation from Pseudo Old Data}: When retaining old training data is impractical, methods like L\&R \cite{xia2022learn}, Wang et al. \cite{wang2022few}, DnR \cite{sun2020distill}, PCLL \cite{zhao2022prompt}, and LFPT5 \cite{qin2021lfpt5} generate synthetic old data. These methods create pseudo-samples that simulate old data distribution. This category is often utilized in generation tasks and named entity recognition.
\end{itemize}

\subsection{Benchmarks and Datasets}

We summarize the commonly used benchmarks and datasets as follows:
(1) \textbf{Continual Text Classification}: CLINC150 \cite{larson2019evaluation}, BANKING77 \cite{casanueva2020efficient}, AGNews, Yelp, Amazon, DBPedia, Yahoo \cite{zhang2015character}, HWU64 \cite{liu2021benchmarking}, (HL5Domains, Liu3Domains, Ding9Domains, SemEval14) \cite{ke2021adapting}, GLUE \cite{wang2018glue};
(2) \textbf{Continual Named Entity Recognition}: OntoNotes5 \cite{hovy2006ontonotes}, I2B2 \cite{murphy2010serving}, Few-NERD \cite{ding2021few};
(3) \textbf{Continual Relation Extraction}: FewRel \cite{han2018fewrel}, TRACRED \cite{zhang2017position};
(4) \textbf{Continual Machine Translation}: WMT \footnote{https://www.statmt.org/}, TED Talks \footnote{http://www.cs.jhu.edu/~kevinduh/a/multitarget-tedtalks/};
(5) \textbf{Continual Knowledge Editing}: zsRE \cite{de2021editing}, FEVER \cite{thorne2018fever}, CounterFact \cite{meng2022locating};
(6) \textbf{Continual Instruction Tuning}: (MNLI,
QQP, RTE, SST2) GLUE \cite{wang2018glue}, (WiC, CB, COPA, MultiRC, BoolQ) SuperGLUE \cite{wang2019superglue}, NaturalInstruction\cite{mishra2022cross}, SuperNI\cite{wang2022super};
(7) \textbf{Continual Alignment}: HH-RLHF \cite{stiennon2020learning}, Reddit TL;DR \cite{volske2017tl};

\section{Methodology: Continual Pretraining}
\label{sec:continual_pretraining}

Continual pretraining \cite{stoppretraining,Ecomgpt,ContinualPre-TrainingofLarge,Continualpre-trainingmitigates,Lifelongpretraining,cert,econet,efficient,elle,examining,investigating,quert,ke2023continual} enhances the internal knowledge of LLMs and is particularly valuable given the high costs associated with full pretraining. Although research on continual pretraining is less developed compared to continual finetuning, it is crucial for enhancing the general capabilities of existing LLMs. There are three types of continual pretraining: \emph{Continual Vertical Domain Pretraining} \cite{ContinualPre-TrainingofLarge,Continualpre-trainingmitigates,corpusbrain,Ecomgpt,efficient,examining,investigating,quert,Recyclable,cert,elle,Lifelongpretraining}, targeting domain-specific continuous learning without forgetting previously acquired expertise; \emph{Continual Language Domain Pretraining} \cite{continuallearning,embracing,Exploringcontinual,Lifelonglanguagepretraining,winata2023overcoming,parameter,simple}, focusing on adapting to evolving language usage; and \emph{Continual Temporal Domain Pretraining} \cite{econet,mitigating,temporalwiki,set,mind,timewaits,timelms}, which updates models with time-sensitive data and enables model to grasp the latest knowledge.

\subsection{Continual Vertical Domain Pretraining}
\label{sec:continual_vertical_domain_pretraining}

Continual Vertical Domain Pretraining \cite{ContinualPre-TrainingofLarge,Continualpre-trainingmitigates,corpusbrain,Ecomgpt,efficient,examining,investigating,quert,Recyclable,cert,elle,Lifelongpretraining} involves continuously training a language model on a series of domain-specific datasets. This method ensures the model performs efficiently across multiple vertical domains or tasks while retaining previously acquired knowledge. For instance, continual pretraining on financial domain data enables LLMs to provide a better analysis of financial texts and data \cite{yang2023fingpt}.

\textbf{Experimental investigations} in continual vertical domain pretraining primarily focus on addressing catastrophic forgetting \cite{investigating,Analyzing,take}. As a pioneering work, Jin et al. \cite{Lifelongpretraining} revealed that distillation-based approaches are most effective in retaining downstream performance in earlier domains. Building on this, Mehta et al. \cite{Anempirical} found that models pre-trained on a diverse set of tasks tend to experience less forgetting compared to those trained from scratch, highlighting the benefits of task diversity. Similarly, Cossu et al. \cite{Continualpre-trainingmitigates} demonstrated that continual pretraining can help mitigate forgetting, supporting the notion that sustained exposure to various tasks can enhance model robustness. However, Li et al. \cite{examining} emphasized that catastrophic forgetting remains a significant challenge and cannot be fully resolved through straightforward methods such as freezing layers, modules, LoRA, and (IA)$^3$ \cite{liu2022few}. These findings collectively underscore the complexity of addressing catastrophic forgetting and the need for innovative approaches in continual vertical domain pretraining. Research on continual vertical domain pretraining has been evolving with various techniques, including but not limited to \emph{experience replay} \cite{domain,elle,Lifelongpretraining,corpusbrain}, \emph{parameter-efficient finetuning} \cite{rho,corpusbrain,large,Lifelongpretraining}, \emph{mixture of experts} \cite{demix,exploringthebenefits}, \emph{knowledge distillation} \cite{Recyclable,Lifelongpretraining}, \emph{model expansion} \cite{elle,bert2bert,llama,solar}, \emph{re-warming} \cite{ContinualPre-TrainingofLarge}, and \emph{data selection} \cite{rho,Ecomgpt,albalak2024survey}.

\subsubsection{\textbf{Parameter-Efficient Fine-Tuning}}
Parameter Efficient Fine-Tuning is a technique designed to optimize models for specific tasks without requiring extensive computational resources.
CorpusBrain++ \cite{corpusbrain} addresses the dynamic nature of real-world knowledge-intensive language tasks by employing a backbone-adapter architecture and an experience replay strategy. In a similar vein, Med-PaLM \cite{large} introduces instruction prompt tuning to the medical domain using a few exemplars. These methods underscore the importance of efficient fine-tuning strategies in adapting LLMs to specialized domains while addressing the challenges of maintaining performance across diverse tasks.

\subsubsection{\textbf{Model Expansion}} \label{sec:continual_vertical_domain_pretraining_model_expansion}
Model expansion involves enhancing the architecture of pre-trained language models by increasing their width and depth to improve efficiency in knowledge acquisition and integration from continuous data streams across multiple domains. ELLE \cite{elle} employs a function-preserved model expansion strategy to achieve this, flexibly expanding the width and depth of existing pre-trained language models. Similarly, bert2BERT \cite{bert2bert} enhances a base BERT model by expanding its architecture, enabling it to better handle new and more complex data while retaining knowledge from earlier training phases. In line with these approaches, LLaMA Pro \cite{llama} expands Transformer blocks and fine-tunes them using a new corpus, achieving superior performance in tasks related to general use, programming, and mathematics. Additionally, SOLAR \cite{solar} utilizes depth up-scaling, which involves depthwise scaling and continued pretraining, to efficiently boost LLM performance across various NLP tasks without necessitating complex changes for training and inference.

\subsubsection{\textbf{Re-warming}}
Re-warming involves adjusting the \emph{learning rate} upwards when introducing new datasets for continual training. Gupta et al. \cite{ContinualPre-TrainingofLarge} propose this strategy to prevent the learning rate from diminishing too much over extended training periods, which can otherwise stall the learning process when new data is introduced. Experimental results show that re-warming the model not only helps in adapting to new datasets more effectively but also enhances overall downstream task performance.

\subsubsection{\textbf{Data Selection}} \label{sec:continual_vertical_domain_pretraining_data_selection}
Data Selection plays a crucial role in pretraining, where various lightweight filters are employed to ensure data quality \cite{rho,albalak2024survey}. These filters include heuristic-based methods (e.g., language and item count filtering), classifier-based methods \cite{brown2020language}, and perplexity-based techniques \cite{wenzek2020ccnet}. For instance, the RedPajama-Data-v2 dataset \cite{together2023redpajama} employs over 40 quality indicators for data filtering and reweighting to enhance data selection.

Recently, Lin et al. \cite{rho} introduced RHO-1, which is trained with Selective Language Modeling (SLM). SLM identifies and prioritizes the most impactful tokens during the training process by assessing the gradient impact of each token, thus giving priority to those that cause higher changes in the loss function. In another approach, LESS \cite{xia2024less} proposes a low-rank gradient similarity search algorithm to efficiently select the most relevant data for targeted instruction tuning, significantly boosting model performance by training on a carefully chosen subset of the data. Additionally, Ma et al. \cite{Ecomgpt} propose EcomGPT-CT, which leverages semi-structured e-commerce data to enhance the model's performance on domain-specific tasks. EcomGPT-CT utilizes a data mixing strategy, integrating general pretraining data with domain-specific semi-structured data, thereby improving its effectiveness in specific domains.

\subsection{Continual Language Domain Pretraining}
\label{sec:continual_language_domain_pretraining}
Continual Language Domain Pretraining \cite{continuallearning,embracing,Exploringcontinual,Lifelonglanguagepretraining,winata2023overcoming,parameter,simple} extends the concept of pretraining language models to continuously integrate new data and adapt to evolving language domains without forgetting previous knowledge. The studies on continual language domain pretraining focus on natural language \cite{embracing,winata2023overcoming,parameter} and code language \cite{Exploringcontinual,evaluating}.
Studies on continual language domain pretraining mainly focus on techniques such as \emph{experience replay} \cite{continuallearning,simple}, \emph{architecture-based methods} \cite{parameter,Exploringcontinual,Lifelonglanguagepretraining,moduleformer}, and \emph{re-warming} \cite{simple}.

\subsubsection{\textbf{Architecture-Based Methods}}
Architecture-Based Methods offer innovative solutions for enhancing the adaptability and efficiency of LLMs in continual language domain pretraining. Yadav et al. \cite{Exploringcontinual} improve prompt tuning by incorporating a teacher forcing mechanism, creating a pool of prompts that guide model finetuning on new tasks and compelling the model to follow specific pathways during training. Yang et al. \cite{embracing} introduce the CLL-CLIP model, which extends the language understanding of CLIP \cite{Learningtransferablevisual} for continual learning of new languages. They employ Token Embedding Initialization and Regularization to mitigate catastrophic forgetting. CLL-CLIP includes an expandable token embedding layer that dynamically adjusts to accommodate linguistic differences, enabling seamless integration of new tokens.
ModuleFormer \cite{moduleformer} and Lifelong-MoE \cite{Lifelonglanguagepretraining} are both architecture-based methods that utilize MoE to enhance LLM efficiency and adaptability. ModuleFormer leverages modularity by activating specific modules based on input tokens, ensuring targeted processing. Lifelong-MoE dynamically adds model capacity by incorporating new experts with regularized pretraining, achieving superior performance in few-shot and multi-task learning scenarios. These methods collectively demonstrate the potential of architectural innovations in addressing the challenges of continual learning.

\subsubsection{\textbf{Re-warming}}
Re-warming, a strategy involving the temporary increase of the learning rate at the start of training on new data, allows the model to adapt more rapidly to new language. Ibrahim et al. \cite{simple} present a continual pretraining approach that combines learning rate (LR) re-warming, LR re-decaying, and data replay. In their method, LR re-warming is followed by LR re-decaying, a systematic reduction of the learning rate according to a specific schedule. This re-decaying phase helps the model stabilize after learning new language, preventing it from overfitting to recent data. This approach aligns with other methods in the field, such as those proposed by Gupta et al. \cite{ContinualPre-TrainingofLarge}, who emphasize the importance of adjusting learning rates to maintain model efficacy during continual vertical domain pretraining.

\subsection{Continual Temporal Domain Pretraining}
\label{sec:continual_temporal_domain_pretraining}

Continual Temporal Domain Pretraining \cite{econet,mitigating,temporalwiki,set,mind,timewaits,timelms} involves continually updating language models with temporally relevant data to maintain their accuracy and relevance as new information becomes available. Existing studies \cite{mind,temporaladaptation,timewaits} highlight that the performance of LLMs degrades over time because they cannot learn new knowledge that is sensitive to temporal changes. For example, an LLM pretrained on 2023 data cannot answer questions about events that happened in 2024.

\textbf{Empirical findings} highlight several challenges in the realm of temporal adaptation for language models. Lazaridou et al. \cite{mind} demonstrate significant performance degradation when models trained on past data are tested on future data, underscoring the struggle of LLMs with temporal generalization. Similarly, Röttger et al. \cite{temporaladaptation} reveal that while temporal adaptation offers slight improvements in masked language model tasks, it does not significantly enhance performance on downstream tasks when compared to domain adaptation alone. Moreover, Luu et al. \cite{timewaits} find that although continual pretraining aids temporal adaptation, it is less effective than task-specific fine-tuning on temporally relevant data, with performance degrading substantially over time. These studies collectively underscore the persistent challenges in achieving robust temporal generalization and the need for more sophisticated adaptation techniques.

Most existing methods utilize experience replay to alleviate forgetting. In addition to experience replay, Han et al. \cite{econet} propose the Effective CONtinual pretraining framework for Event Temporal reasoning (ECONET), which integrates targeted masking and contrastive loss to emphasize event and temporal indicators during training. Specifically, ECONET employs a mask prediction strategy, where specific tokens related to events and times are masked, and a discriminator model is used to distinguish correct from corrupted sentences, thus enhancing temporal reasoning. Zhao et al. \cite{set} introduce temporal-adaptive finetuning, which synchronizes the internal knowledge of the model with a target time without altering the explicit contextual information provided to the model. Complementing these approaches, TimeLMs \cite{timelms} are continually updated language models trained on diachronic Twitter data to capture temporal changes in language and maintain relevance over time. Together, these methods demonstrate innovative strategies for addressing the challenges of continual learning and temporal adaptation in language models.

\subsection{Summary}
Continual pretraining enhances LLMs by updating their internal knowledge without incurring the high costs of full pretraining. Current research spans vertical, language, and temporal domains, addressing challenges like catastrophic forgetting and temporal adaptation. Techniques such as experience replay, knowledge distillation, parameter-efficient finetuning, model expansion, and re-warming have shown promise. Despite these advances, significant challenges remain, particularly in maintaining performance over time and across diverse tasks. Future research should focus on innovative approaches to mitigate forgetting, improve temporal generalization, and develop efficient, adaptive architectures for sustained model performance.

\begin{figure}[!t]
    \centering
    \subfloat[Continual Text Classification]{
        \includegraphics[width=0.47\linewidth]{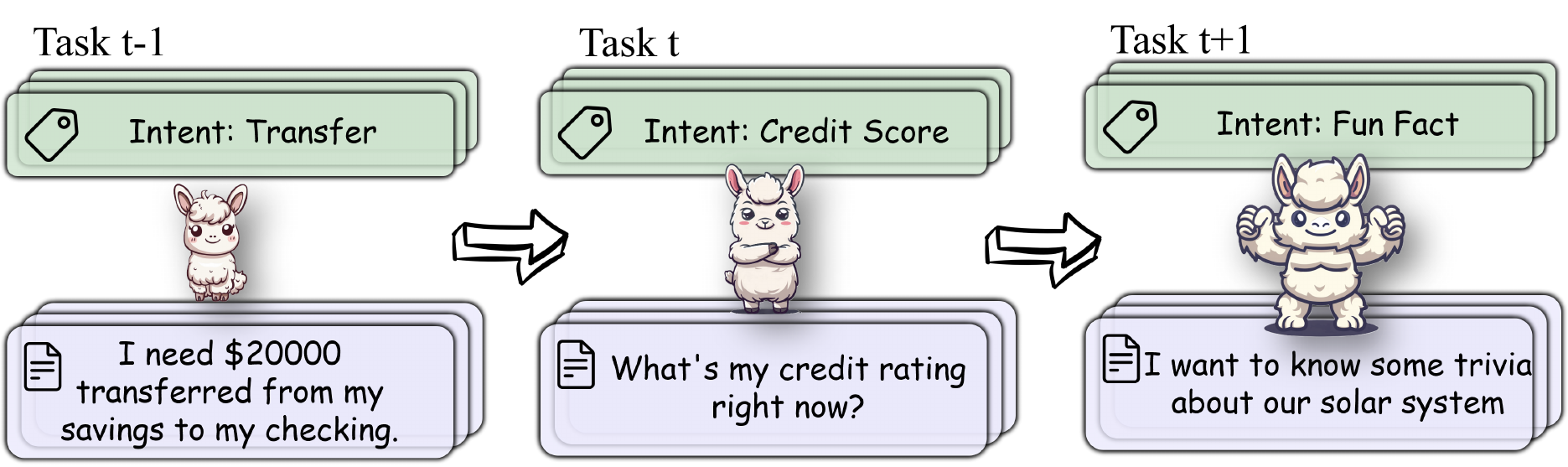}
    }
    \hspace{0.5cm}
    \subfloat[Continual Named Entity Recognition]{
        \includegraphics[width=0.47\linewidth]{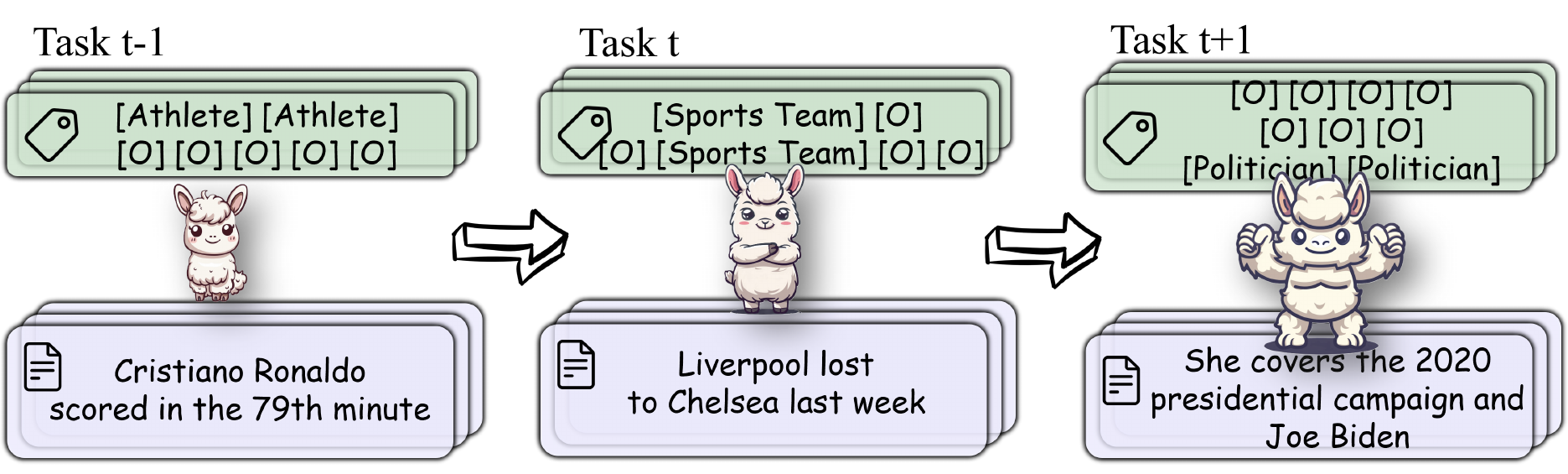}
    }

    \subfloat[Continual Relation Extraction]{
        \includegraphics[width=0.47\linewidth]{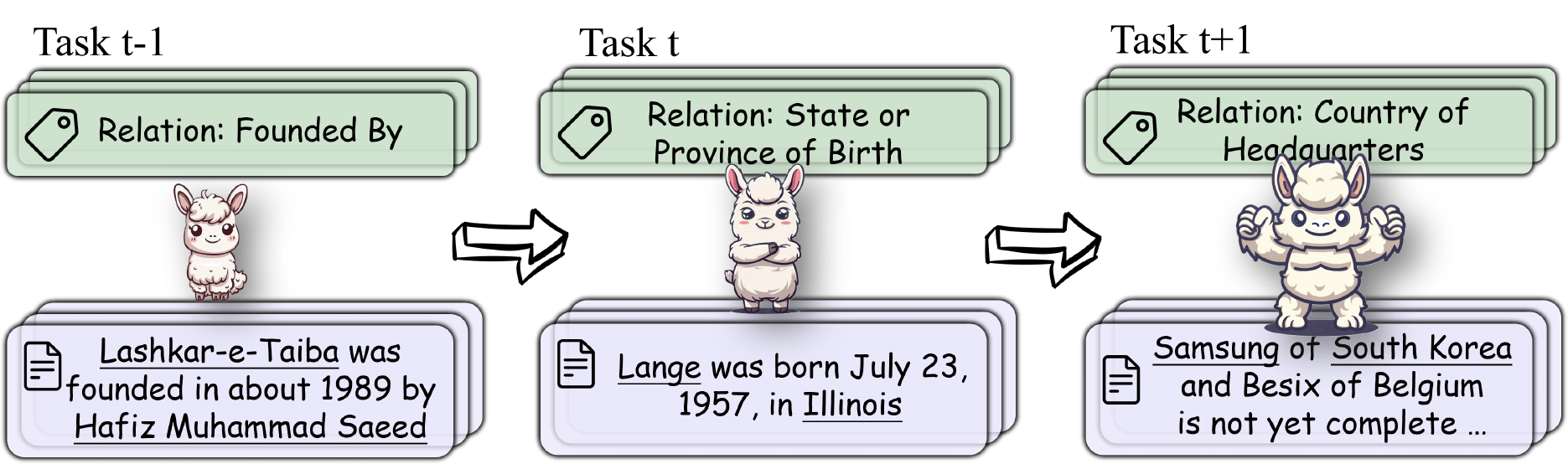}
    }
    \hspace{0.5cm}
    \subfloat[Continual Knowledge Editing]{
        \includegraphics[width=0.47\linewidth]{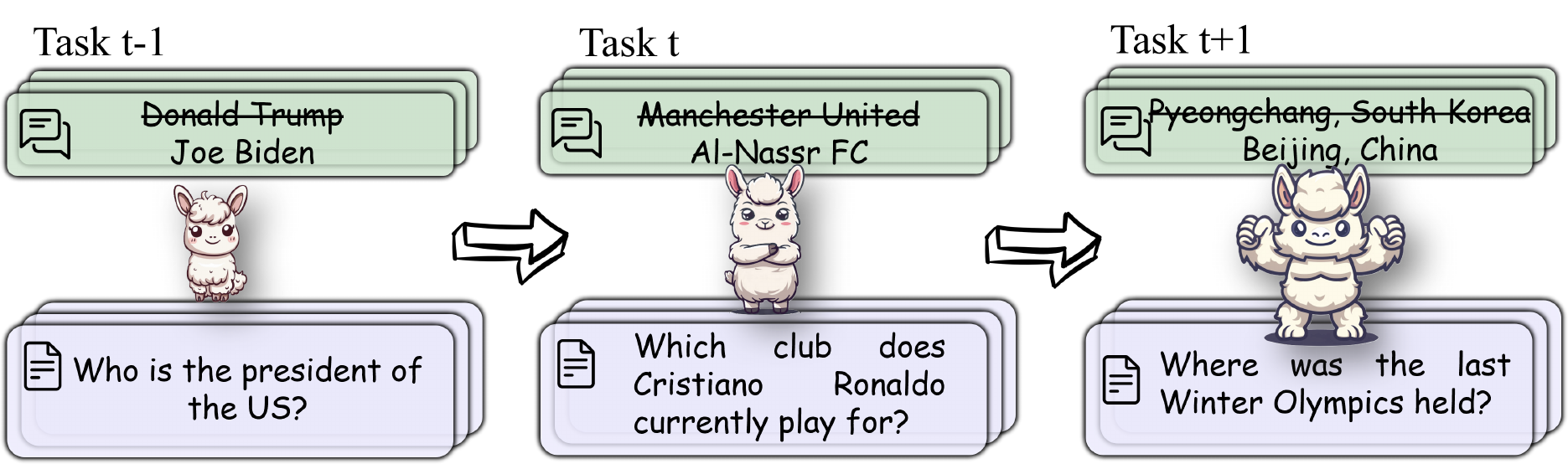}
    }
    
    \subfloat[Continual Machine Translation]{
        \includegraphics[width=0.47\linewidth]{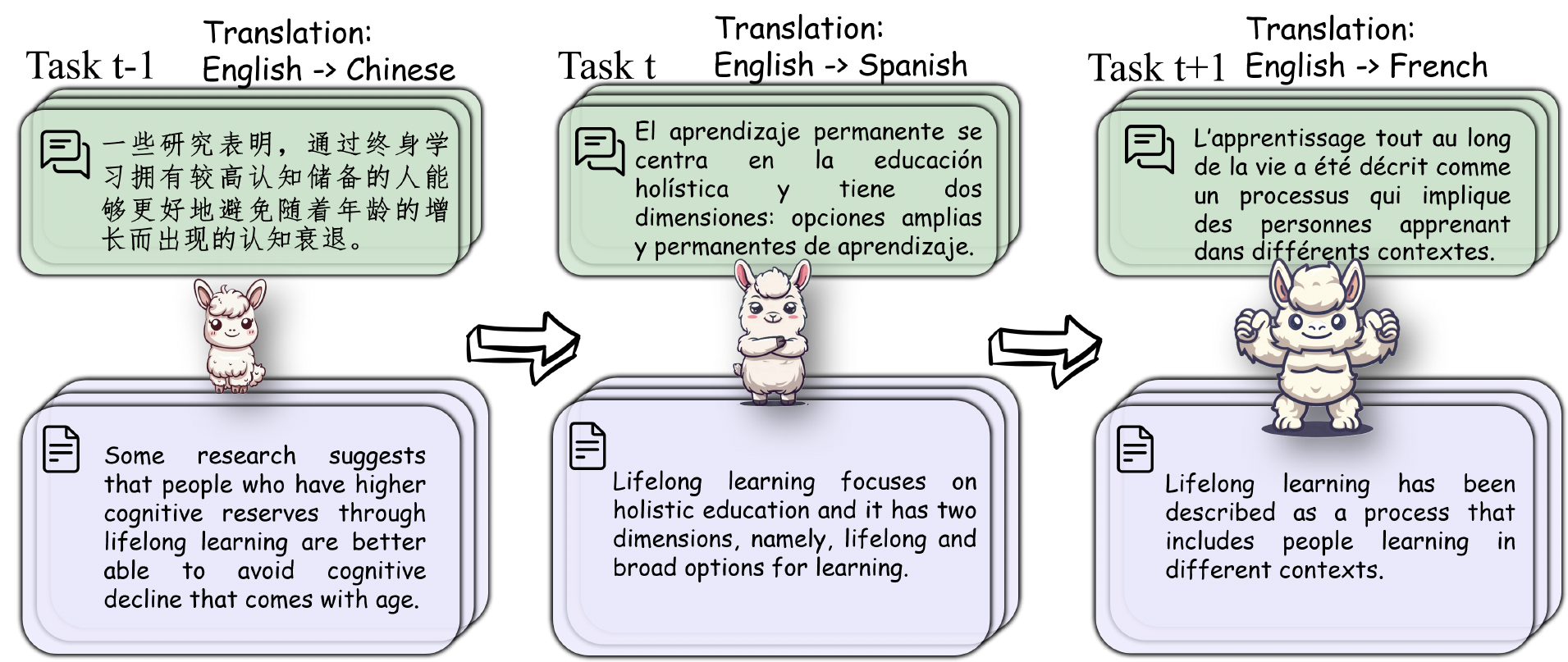}
    }
    \hspace{0.5cm}
    \subfloat[Continual Instruction Tuning]{
        \includegraphics[width=0.47\linewidth]{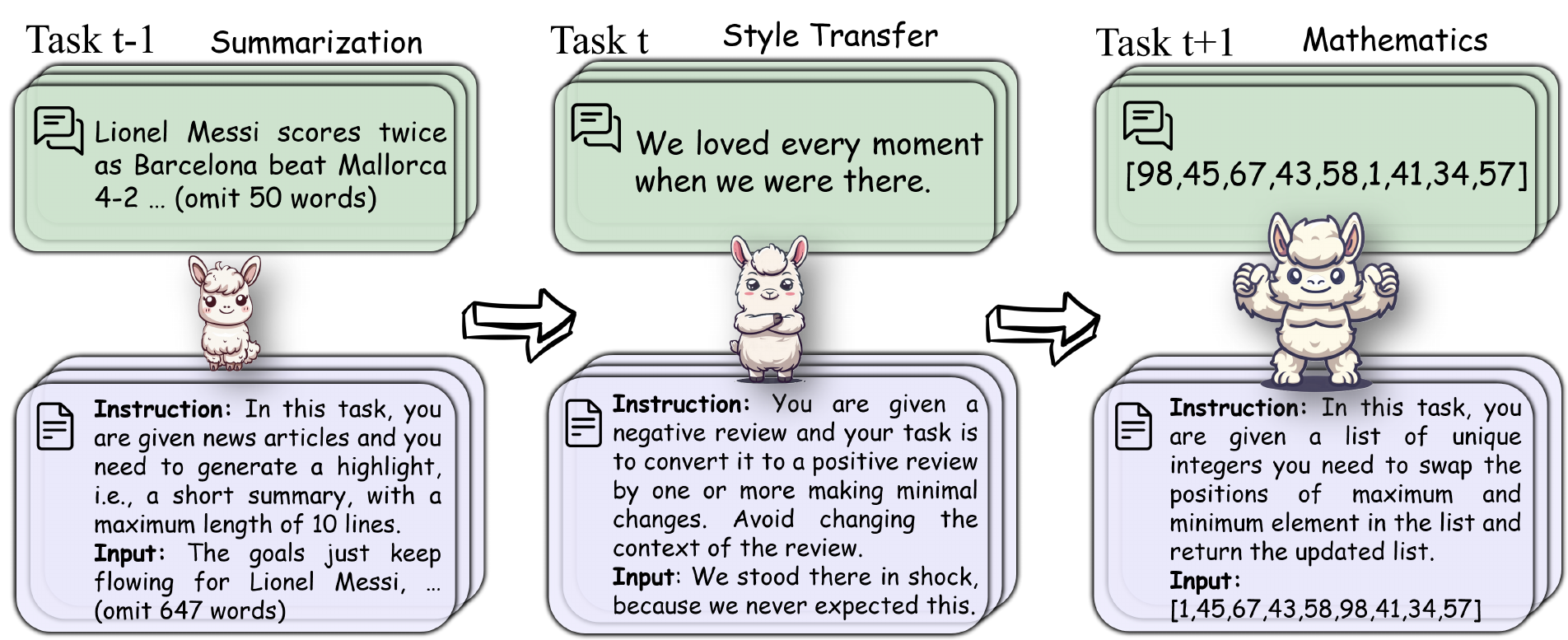}
    }
    
    \subfloat[Continual Alignment]{
        \includegraphics[width=0.55\linewidth]{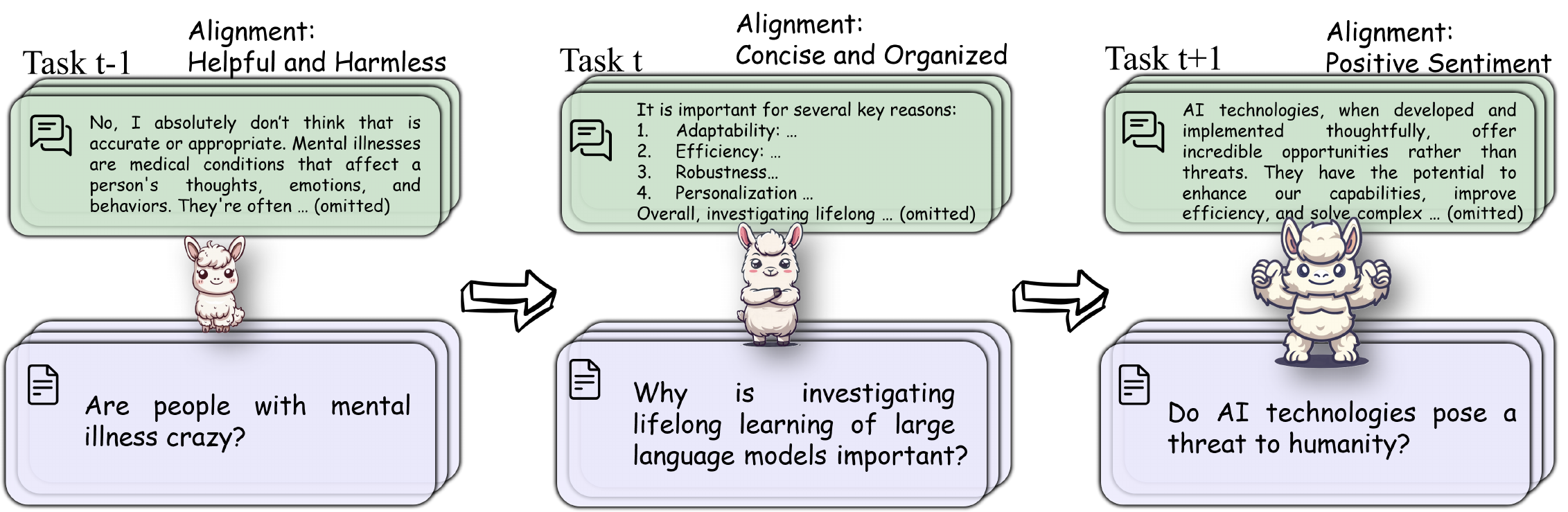}
    }
    \caption{An illustration of continual finetuning scenarios. In each continual finetuning scenario, a model learns task $t-1$, $t$, and $t+1$ sequentially (left to right). The \textbf{\textcolor[RGB]{159,116,191}{PURPLE}} and the \textbf{\textcolor[RGB]{122,142,107}{GREEN}} boxes represent the input and the output respectively.}
    \label{fig:illustration_continual_finetuning}
\end{figure}
\section{Methodology: Continual Finetuning}
\label{sec:continual_finetuning}

\begin{table}[!t]
  \centering
  \caption{Comparison between representative methods for \textbf{continual text classification} and \textbf{continual named entity recognition}. \textbf{PEFT} represents whether utilize parameter-efficient finetuning methods for training models. \textbf{Replay, Regularization, Distillation, Architecture} refer to the common techniques summarized in Section~\ref{sec:overview_common_techniqes}. 
  }
  \resizebox{\linewidth}{!}{
    \begin{tabular}{llllp{6cm}lllllll}
    \toprule
    \multicolumn{1}{c}{\textbf{Method}} & \multicolumn{1}{c}{\textbf{Year}} & \multicolumn{1}{c}{\textbf{Publication}} & \multicolumn{1}{c}{\textbf{Backbone}} & \multicolumn{1}{c}{\textbf{Dataset}} & \multicolumn{1}{c}{\textbf{Code}} & \multicolumn{1}{c}{\textbf{PEFT}} & \multicolumn{1}{c}{\textbf{Replay}} & \multicolumn{1}{c}{\textbf{Distillation}} & \multicolumn{1}{c}{\textbf{Regularization}} & \multicolumn{1}{c}{\textbf{Architecture}} & \multicolumn{1}{c}{\textbf{Others}} \\
    \midrule
    \rowcolor[rgb]{ .788,  .788,  .788} \multicolumn{12}{c}{\textit{\textbf{Continual Text Classification}}} \\
    \midrule
    CL-KD \cite{castellucci2021learning} & 2021  & ACL   & BERT  & PAWS-X, MARC, CoNLL 2002, CoNLL 2003 & /     & /     & /     & \color[RGB]{3,191,61}{\Checkmark} & /     & /     & / \\
    \midrule
    \rowcolor[rgb]{ .867,  .922,  .969} B-CL \cite{ke2021adapting} & 2021  & NAACL & BERT  & HL5Domains, Liu3Domains, Ding9Domains, SemEval14 & \href{https://github.com/ZixuanKe/PyContinual}{Link} & Capsule Network & /     & /     & /     & \color[RGB]{3,191,61}{\Checkmark} & / \\
    \midrule
    CLASSIC \cite{ke2021classic} & 2021  & EMNLP & BERT  & HL5Domains, Liu3Domains, Ding9Domains, SemEval14 & \href{https://github.com/ZixuanKe/PyContinual}{Link} & Adapters & /     & /     & /     & \color[RGB]{3,191,61}{\Checkmark} & / \\
    \midrule
    \rowcolor[rgb]{ .867,  .922,  .969} IDBR \cite{huang2021continual} & 2021  & NAACL & BERT  & AGNews, Yelp, Amazon, DBPedia, Yahoo & \href{https://github.com/GT-SALT/IDBR}{Link} & /     & \color[RGB]{3,191,61}{\Checkmark} & /     & \color[RGB]{3,191,61}{\Checkmark} & /     & / \\
    \midrule
    CCFI \cite{hua2021hyperparameter} & 2021  & NAACL & BERT  & CLINC150 & \href{https://github.com/tinghua-code/CCFI}{Link} & /     & \color[RGB]{3,191,61}{\Checkmark} & /     & \color[RGB]{3,191,61}{\Checkmark} & /     & / \\
    \midrule
    \rowcolor[rgb]{ .867,  .922,  .969} ENTAILMENT \cite{xia2021incremental} & 2021  & NAACL & RoBERTa & Banking77, FewRel & \href{https://github.com/congyingxia/IncrementalFSTC}{Link} & /     & /     & /     & /     & /     & / \\
    \midrule
    CTR \cite{ke2021achieving} & 2021  & NIPS  & BERT  & HL5Domains, Liu3Domains, Ding9Domains, SemEval14 & \href{https://github.com/ZixuanKe/PyContinual}{Link} & Capsule Network & /     & /     & /     & /     & / \\
    \midrule
    \rowcolor[rgb]{ .867,  .922,  .969} IPRLS \cite{geng2021iterative} & 2021  & SIGIR & BERT  & Amazon, IMDB, MR & \href{https://github.com/siat-nlp/IPRLS}{Link} & /     & /     & /     & \color[RGB]{3,191,61}{\Checkmark} & /     & Pruning \\
    \midrule
    MSR \cite{liu2021lifelong} & 2021  & /     & BERT  & ATIS, SNIPS, HWU64, CLINC150 & /     & /     & \color[RGB]{3,191,61}{\Checkmark} & \color[RGB]{3,191,61}{\Checkmark} & \color[RGB]{3,191,61}{\Checkmark} & /     & / \\
    \midrule
    \rowcolor[rgb]{ .867,  .922,  .969} DR-EMR \cite{vijayaraghavan2021lifelong} & 2021  & EACL  & BERT  & ATOMIC, CONCEPTNET, SB-SCK & \href{https://pralav.github.io/lifelong_eventrep/?c=10}{Link} & /     & \color[RGB]{3,191,61}{\Checkmark} & /     & \color[RGB]{3,191,61}{\Checkmark} & /     & / \\
    \midrule
    Qian et. al. \cite{qian2021lifelong} & 2021  & NAACL & BERT  & SPLC  & /     & /     & \color[RGB]{3,191,61}{\Checkmark} & /     & \color[RGB]{3,191,61}{\Checkmark} & /     & / \\
    \midrule
    \rowcolor[rgb]{ .867,  .922,  .969} PLE \cite{li2022continual} & 2022  & COLING & RoBERTa & CLINC150, ATIS, HWU64, BANKING77, MTOP, SNIPS, LEYZER, MSLU, TOP & /     & Prefix Tuning, Adapters & Pseudo Sample & \color[RGB]{3,191,61}{\Checkmark} & /     & /     & / \\
    \midrule
    CRN \cite{bai2022incremental} & 2022  & ACL (Findings) & BERT  & KUAKE-QIC, CMID & /     & /     & \color[RGB]{3,191,61}{\Checkmark} & \color[RGB]{3,191,61}{\Checkmark} & /     & /     & Contrastive Learning \\
    \midrule
    \rowcolor[rgb]{ .867,  .922,  .969} CPT \cite{ke2022continual} & 2022  & EMNLP & RoBERTa & AGNews, ACL-ARC, SCIERC, SemEval-res & \href{https://github.com/UIC-Liu-Lab/CPT}{Link} & Adapters & /     & /     & /     & \color[RGB]{3,191,61}{\Checkmark} & / \\
    \midrule
    PE \cite{zhu2022parameter} & 2022  & NAACL & BERT  & Amazon Reviews & /     & Parameter Selection & /     & \color[RGB]{3,191,61}{\Checkmark} & \color[RGB]{3,191,61}{\Checkmark} & /     & / \\
    \midrule
    \rowcolor[rgb]{ .867,  .922,  .969} PAGeR \cite{varshney2022prompt} & 2022  & NAACL (Findings) & GPT-2 & CLINC150, BANKING77, HWU64, SGD, Stackoverflow, MWOZ & /     & /     & \color[RGB]{3,191,61}{\Checkmark} & \color[RGB]{3,191,61}{\Checkmark} & /     & /     & Contrastive Learning \\
    \midrule
    ADA \cite{ermis2022memory} & 2022  & NIPS  & BERT, DistilBERT, RoBERTa & Arxiv-Papers, Reuter, Wiki-30K & /     & Adapters & /     & \color[RGB]{3,191,61}{\Checkmark} & /     & /     & / \\
    \midrule
    \rowcolor[rgb]{ .867,  .922,  .969} SCCL \cite{luo2023mitigating} & 2023  & /     & RoBERTa & CoLA, MNLI, QNLI, QQP, Yelp, AGNews & /     & /     & \color[RGB]{3,191,61}{\Checkmark} & \color[RGB]{3,191,61}{\Checkmark} & /     & /     & Contrastive Learning \\
    \midrule
    SEQ* \cite{zheng2023learn} & 2023  & /     & BERT, GPT2, Pythia & CLINC150, BANKING77, AGNews, Yelp, Amazon, DBPedia, Yahoo & \href{https://github.com/zzz47zzz/pretrained-lm-for-incremental-learning}{Link} & /     & /     & /     & /     & /     & Classifier Expasion \\
    \midrule
    \rowcolor[rgb]{ .867,  .922,  .969} EPI \cite{wang2023rehearsal} & 2023  & ACL   & BERT  & AGNews, Yelp, Amazon, DBPedia, Yahoo, WOS & \href{https://github.com/Dicer-Zz/EPI}{Link} & Prefix Tuning & /     & /     & /     & \color[RGB]{3,191,61}{\Checkmark} & / \\
    \midrule
    VAG \cite{shao2023class} & 2023  & ACL   & BART  & CLINC150, BANKING77, 20 Newsgroups, FewRel, TACRED & \href{https://github.com/shaoyijia/VAG}{Link} & /     & Label-based Pseudo Replay & /     & /     & /     & Vocabulary \\
    \midrule
    \rowcolor[rgb]{ .867,  .922,  .969} LR ADJUST \cite{winata2023overcoming} & 2023  & ACL (Findings) & XLM-R & MASSIVE, WikiAnn & /     & /     & /     & /     & /     & /     & Adjusts Learning Rate \\
    \midrule
    InfoCL \cite{song2023infocl} & 2023  & EMNLP & BERT  & HWU64, FewRel, TACRED, MAVEN,  & \href{https://github.com/Yifan-Song793/InfoCL}{Link} & /     & \color[RGB]{3,191,61}{\Checkmark} & \color[RGB]{3,191,61}{\Checkmark} & /     & /     & Contrastive Learning \\
    \midrule
    \rowcolor[rgb]{ .867,  .922,  .969} HOP \cite{michieli2024hop} & 2024  & /     & BERT  & HL5Domains, Liu3Domains, Ding9Domains, SemEval14, NLI, 20News, DSC & /     & Adapters & /     & /     & /     & \color[RGB]{3,191,61}{\Checkmark} & / \\
    \midrule
    EKFAC \cite{chen2024bayesian} & 2024  & /     & OPT   & MNLI, QQP, QNLI, SST-2 & \href{https://recherchetts.github.io/bayesian-peft/}{Link} & LoRA  & /     & /     & \color[RGB]{3,191,61}{\Checkmark} & /     & / \\
    \midrule
    \rowcolor[rgb]{ .867,  .922,  .969} MoCL \cite{wang2024rehearsal} & 2024  & NAACL & BERT, T5, LLaMA & WOS, AGNews, Yelp, Amazon, DBPedia, Yahoo & \href{https://github.com/boschresearch/MoCL-NAACL-2024}{Link} & LoRA, Prefix Tuning & /     & /     & /     & \color[RGB]{3,191,61}{\Checkmark} & / \\
    \midrule
    \rowcolor[rgb]{ .788,  .788,  .788} \multicolumn{12}{c}{\textit{\textbf{Continual Named Entity Recognition}}} \\
    \midrule
    ProgModel \cite{shen2019progressive} & 2019  & EMNLP & RNN   & ATIS, Snips & /     & /     & /     & /     & /     & \color[RGB]{3,191,61}{\Checkmark} & / \\
    \midrule
    \rowcolor[rgb]{ .988,  .894,  .839} KCN \cite{cao2020incremental} & 2020  & EMNLP & BERT  & ACE 2005, TAC KBP 2017 & \href{https://github.com/CPF-NLPR/IncrementalED}{Link} & /     & \color[RGB]{3,191,61}{\Checkmark} & \color[RGB]{3,191,61}{\Checkmark} & /     & /     & / \\
    \midrule
    ExtendNER, AddNER \cite{monaikul2021continual} & 2021  & AAAI  & BERT  & CoNLL 2003, OntoNotes5 & /     & /     & /     & \color[RGB]{3,191,61}{\Checkmark} & /     & /     & / \\
    \midrule
    \rowcolor[rgb]{ .988,  .894,  .839} KD+R+K \cite{yu2021lifelong} & 2021  & EMNLP & BERT  & ACE 2005, MAVEN & \href{https://github.com/Perfec-Yu/Lifelong-ED}{Link} & /     & \color[RGB]{3,191,61}{\Checkmark} & \color[RGB]{3,191,61}{\Checkmark} & /     & /     & / \\
    \midrule
    Wang et. al. \cite{wang2022few} & 2022  & ACL   & BERT  & CoNLL 2003, OntoNotes5 & /     & /     & Pseudo Sample & \color[RGB]{3,191,61}{\Checkmark} & /     & /     & / \\
    \midrule
    \rowcolor[rgb]{ .988,  .894,  .839} L\&R \cite{xia2022learn} & 2022  & ACL (Findings) & BERT  & CoNLL 2003, OntoNotes5 & /     & /     & Pseudo Sample & \color[RGB]{3,191,61}{\Checkmark} & /     & /     & / \\
    \midrule
    EMP \cite{liu2022incremental} & 2022  & COLING & BERT  & ACE 2005, MAVEN & \href{https://github.com/VT-NLP/Incremental_Prompting}{Link} & /     & \color[RGB]{3,191,61}{\Checkmark} & \color[RGB]{3,191,61}{\Checkmark} & /     & /     & / \\
    \midrule
    \rowcolor[rgb]{ .988,  .894,  .839} CFNER \cite{zheng2022distilling} & 2022  & EMNLP & BERT  & CoNLL 2003, OntoNotes5, I2B2 & \href{https://github.com/zzz47zzz/CFNER}{Link} & /     & /     & \color[RGB]{3,191,61}{\Checkmark} & /     & /     & Causal Effect \\
    \midrule
    BNU \cite{li2022bnu} & 2022  & ICASSP & BERT  & ACE 2005, TAC KBP 2017 & /     & /     & \color[RGB]{3,191,61}{\Checkmark} & \color[RGB]{3,191,61}{\Checkmark} & \color[RGB]{3,191,61}{\Checkmark} & /     & / \\
    \midrule
    \rowcolor[rgb]{ .988,  .894,  .839} SDAPN \cite{chen2022similarity} & 2022  & ICTAI & BERT  & CoNLL 2003, OntoNotes5 & /     & /     & \color[RGB]{3,191,61}{\Checkmark} & \color[RGB]{3,191,61}{\Checkmark} & /     & /     & Prototype \\
    \midrule
    HEFT \cite{wei2022heft} & 2022  & KBS   & BERT  & ACE 2005, TAC KBP 2017 & /     & /     & \color[RGB]{3,191,61}{\Checkmark} & \color[RGB]{3,191,61}{\Checkmark} & \color[RGB]{3,191,61}{\Checkmark} & /     & / \\
    \midrule
    \rowcolor[rgb]{ .988,  .894,  .839} ConPET \cite{song2023conpet} & 2023  & /     & LLaMA & OntoNotes5, Few-NERD, BBN, ACE 2005 & \href{https://github.com/Raincleared-Song/ConPET}{Link} & LoRA  & /     & /     & /     & \color[RGB]{3,191,61}{\Checkmark} & / \\
    \midrule
    SEQ* \cite{zheng2023learn} & 2023  & /     & BERT, GPT2, Pythia & OntoNotes5, I2B2, Few-NERD & \href{https://github.com/zzz47zzz/pretrained-lm-for-incremental-learning}{Link} & /     & /     & /     & /     & /     & Classifier Expasion \\
    \midrule
    \rowcolor[rgb]{ .988,  .894,  .839} SpanKL \cite{zhang2023neural} & 2023  & AAAI  & BERT  & OntoNotes5, Few-NERD & \href{https://github.com/Qznan/SpanK}{Link} & /     & /     & \color[RGB]{3,191,61}{\Checkmark} & /     & /     & Span-Level Prediction \\
    \midrule
    OCILNER \cite{ma2023learning} & 2023  & ACL   & BERT  & CoNLL 2003, OntoNotes5, Few-NERD & \href{https://github.com/rtmaww/O_CILNER}{Link} & /     & \color[RGB]{3,191,61}{\Checkmark} & /     & /     & /     & Contrastive Learning, Prototype \\
    \midrule
    \rowcolor[rgb]{ .988,  .894,  .839} ICE \cite{liu2023teamwork} & 2023  & ACL Findings & BERT  & Few-NERD, MAVEN, ACE 2005 & \href{https://github.com/VT-NLP/ICE}{Link} & /     & /     & /     & /     & \color[RGB]{3,191,61}{\Checkmark} & Frozen Backbones \\
    \midrule
    ProtoNER \cite{kumar2023protoner} & 2023  & BPM   & LayoutLMv2 & Purchase Order & /     & /     & /     & \color[RGB]{3,191,61}{\Checkmark} & /     & /     & Prototype \\
    \midrule
    \rowcolor[rgb]{ .988,  .894,  .839} RDP \cite{zhang2023task} & 2023  & CIKM  & BERT  & CoNLL 2003, OntoNotes5, I2B2 & \href{https://github.com/BladeDancer957/INER_RDP}{Link} & /     & /     & \color[RGB]{3,191,61}{\Checkmark} & /     & /     & Prototype \\
    \midrule
    CPFD \cite{zhang2023continual} & 2023  & EMNLP & BERT  & CoNLL 2003, OntoNotes5, I2B2 & \href{https://github.com/BladeDancer957/CPFD}{Link} & /     & /     & \color[RGB]{3,191,61}{\Checkmark} & \color[RGB]{3,191,61}{\Checkmark} & /     & / \\
    \midrule
    \rowcolor[rgb]{ .988,  .894,  .839} SKD-NER \cite{chen2023skd} & 2023  & EMNLP & BERT  & OntoNotes5, Few-NERD & /     & /     & /     & \color[RGB]{3,191,61}{\Checkmark} & /     & /     & Reinforcement Learning \\
    \midrule
    Liang et. al. \cite{liang2023novel} & 2023  & EMNLP (Findings) & BERT  & ATIS, Snips & \href{https://github.com/cs-liangchen-work/NovelIE}{Link} & /     & \color[RGB]{3,191,61}{\Checkmark} & \color[RGB]{3,191,61}{\Checkmark} & /     & /     & Prototype \\
    \midrule
    \rowcolor[rgb]{ .988,  .894,  .839} Lin et. al. \cite{lin2023incremental} & 2023  & Neurocomputing & BERT  & ACE 2005, MAVEN & /     & /     & /     & \color[RGB]{3,191,61}{\Checkmark} & /     & /     & / \\
    \midrule
    DLD \cite{zhang2023decomposing} & 2023  & SIGIR & BERT  & CoNLL 2003, OntoNotes5, I2B2 & /     & /     & /     & \color[RGB]{3,191,61}{\Checkmark} & /     & /     & / \\
    \midrule
    \rowcolor[rgb]{ .988,  .894,  .839} IS3 \cite{qiu2024incremental} & 2024  & /     & BERT  & OntoNotes5, I2B2, MAVEN & /     & /     & /     & \color[RGB]{3,191,61}{\Checkmark} & /     & /     & / \\
    \midrule
    IFSED \cite{wang2024few} & 2024  & TALLIP & BERT  & FewEvent & /     & /     & \color[RGB]{3,191,61}{\Checkmark} & \color[RGB]{3,191,61}{\Checkmark} & \color[RGB]{3,191,61}{\Checkmark} & /     & Prototype \\
    \bottomrule
\end{tabular}%
    }
  \label{tab:TC_NER_methods}%
\end{table}%

Continual finetuning \cite{huang2021continual,monaikul2021continual,wang2019sentence,sun2019lamol,lin2023mitigating,CKEsurvey} enhances the internal knowledge of LLMs and adapts LLMs to specific tasks such as text classification \cite{huang2021continual}, named entity recognition \cite{monaikul2021continual}, relation extraction \cite{wang2019sentence}, machine translation \cite{cao2021continual} or general generation tasks such as instruction tuning \cite{sun2019lamol}, knowledge editing \cite{CKEsurvey}, and alignment with human preference \cite{lin2023mitigating}.
We provide an illustration of the 7 continual finetuning scenarios in Figure~\ref{fig:illustration_continual_finetuning}.

\subsection{Continual Text Classification}
\label{sec:continual_text_classification}

Text classification includes different directions, such as intent detection, sentiment classification, topic classification, and domain classification. However, past text classification methods can only detect predefined categories. In the real world, new categories may constantly challenge the deployed models. For example, the COVID-19 pandemic brought many new topic categories such as ``nucleic acid detection'' and ``group immunity''. Thus, the emergence of Continual Text Classification allows models to continuously learn new data and recognize new emerging categories. The methods can be broadly divided into the following main categories: \emph{distillation-based} \cite{ke2021classic,liu2021lifelong}, \emph{replay-based} \cite{vijayaraghavan2021lifelong,varshney2022prompt,bai2022incremental,liu2023class,song2023infocl,li2022continual}, \emph{regularization-based} \cite{zhu2022parameter,huang2021continual,hua2021hyperparameter,qian2021lifelong,chen2024bayesian}, \emph{architecture-based}, and others \cite{pasunuru2021continual,ke2021adapting,xia2021incremental,castellucci2021learning,winata2023overcoming}. A detailed comparison between these methods is provided in Table~\ref{tab:TC_NER_methods}.

\subsubsection{\textbf{Distillation-Based}}
To enhance the discriminability between text categories, CLASSIC \cite{ke2021classic} employs contrastive ensemble distillation, enhancing knowledge transfer across tasks through contrastive losses. In addition this, multi-strategy rebalancing, combining cosine normalization, hierarchical knowledge distillation, and interclass margin loss are introduced by MSR \cite{liu2021lifelong} to tackle class imbalance.

\subsubsection{\textbf{Replay-Based}}
Several approaches integrate contrastive learning techniques or structured learning methods to enhance the quality of replay samples and the stability of the learning process. SCN \cite{liu2023class} and InfoCL \cite{song2023infocl} optimize sample selection and leverage contrastive learning for better representation recovery and to combat replay overfitting. These methods help maintain coherence and relevance of the learned representations, addressing issues like data imbalance and the presence of rare words in specific domains.

Each method incorporates adaptations tailored to specific domains or tasks, ensuring that the continual learning model effectively handles the unique challenges and characteristics of those domains. For example, DR-EMR \cite{vijayaraghavan2021lifelong} integrates social commonsense knowledge, and CRN \cite{bai2022incremental} specifically targets the medical field's challenges, showing a deep integration of domain-specific knowledge into the learning processes.

Innovative memory management strategies such as DR-EMR \cite{vijayaraghavan2021lifelong}, PAGeR \cite{varshney2022prompt} and the use of lightweight encoders PLE \cite{li2022continual} with prefix guidance are employed to mitigate catastrophic forgetting and promote efficient knowledge retention. These strategies include selecting representative samples that best capture the essence of previous tasks and employing lightweight models that adapt more dynamically to new information without losing previous knowledge.

\subsubsection{\textbf{Regularization-Based}}
To improve the efficiency of parameter updates, techniques such as selectively updating a small subset of parameters, as seen in the PE \cite{zhu2022parameter}, IDBR \cite{huang2021continual}, and EKFAC \cite{chen2024bayesian} prioritize reducing the computational burden. These methods ensure that the learning process is both resource-efficient and effective at integrating new knowledge without overwriting valuable information from previous tasks.

To automate the adjustment of regularization processes, several approches eliminate the need for manual hyperparameter tuning, allowing the model to adaptively balance retaining old knowledge with acquiring new information, as showcased in CCFI \cite{hua2021hyperparameter}, Qian et al. \cite{qian2021lifelong}.

\subsubsection{\textbf{Architecture-Based}}
To enhance knowledge sharing, there are several approaches to propose relevant strategies, such as hierarchical overlay projections in the HOP \cite{michieli2024hop} and dynamic routing mechanisms in B-CL \cite{ke2021adapting} and CTR \cite{ke2021achieving}, and ADA \cite{ermis2022memory}. These strategies optimize the transfer and sharing of knowledge across different tasks, improving the efficiency and effectiveness of the model when learning new tasks.

To protect task-specific knowledge, several studies introduce mechanisms for parameter isolation, such as B-CL's \cite{ke2021adapting} continual learning adapter, selective activation/deactivation of transformer components, instance-wise relation distillation in SCCL \cite{luo2023mitigating} and private parameter isolation in EPI \cite{wang2023rehearsal}. These approaches effectively minimize interference between new and old tasks, maintaining performance on previous tasks while integrating new ones, thus addressing catastrophic forgetting.

\subsubsection{\textbf{Others}}
In addition to continual text classification tasks, there are also tasks focused on few-shot text classification and multilingual text classification, such as Pasunuru et al. \cite{pasunuru2021continual} and ENTAILMENT \cite{xia2021incremental} focus on improving few-shot learning capabilities, which involve training models with very few examples per class, CL-KD \cite{castellucci2021learning} and LR ADJUST \cite{winata2023overcoming} continually integrate new languages into an existing model, alleviating catastrophic forgetting in multilingual settings.

\subsection{Continual Named Entity Recognition}
\label{sec:continual_named_entity_recognition}

Continual Named Entity Recognition is designed to adaptively identify novel entity types, addressing the dynamic emergence of new entities in the real world. It involves incrementally training models on newly annotated datasets that contain only these novel entities, enabling the models to gradually expand their recognition capabilities to include these new classes without forgetting previously learned entities. For example, in the sentence ``Liverpool lost to Chelsea last week'', a continual named entity recognition model aims to correctly label ``Liverpool'' and ``Chelsea'' as [Sports Team], while non-entity tokens are labeled as [Other]. This approach allows the model to adapt to recognize new entities such as [Politician] in other contexts.

In addition to the challenge of catastrophic forgetting, continual named entity recognition must also contend with semantic shifts \cite{zheng2022distilling,qiu2024incremental}. Semantic shift occurs when the classification of a label changes, for instance, from ``Other'' to a specific entity type, or vice versa. This is particularly challenging as only entities relevant to the current task are annotated, while both previously learned and unseen entities are labeled as ``Other''. 
Existing methods can be broadly classified into four primary categories: \emph{distillation-based} \cite{monaikul2021continual,zhang2023decomposing,zhang2023task,zhang2023continual,zheng2022distilling,zhang2023neural,chen2023skd}, \emph{replay-based} \cite{xia2022learn,ma2023learning,cao2020incremental,yu2021lifelong}, \emph{prototype-based} \cite{chen2022similarity,kumar2023protoner,qiu2024incremental}, \emph{architecture-based} \cite{shen2019progressive,liu2023teamwork,song2023conpet}. 
A detailed comparison between these methods is provided in Table~\ref{tab:TC_NER_methods}.

\subsubsection{\textbf{Distillation-Based}}
In general continual learning scenarios, feature-level knowledge distillation is commonly employed to impose \textit{implicit} knowledge constraints on the student model in the feature space. In continual named entity recognition, knowledge distillation involves inputting new training examples into the teacher model and guiding the student model using the resulting logits. This effectively utilizes old samples from new training examples for \textit{implicit replay}, thereby impose \textit{explicit} knowledge constrains on the student model.
As a pioneer work, ExtendNER \cite{monaikul2021continual}, considering the realistic scenario of continuously emerging named entities, introduces knowledge distillation into named entity recognition to construct a framework for continuous named entity recognition.
Subsequent methods, integrating knowledge distillation techniques, have been improved to address semantic shift caused by the "Other" entity type, such as DLD \cite{zhang2023decomposing}, RDP \cite{zhang2023task}, CPFD \cite{zhang2023continual}, etc. 
In addition to this, some methods have introduced new perspectives or technologies.
CFNER \cite{zheng2022distilling} establishes a causal framework \cite{zheng2023preserving,zheng2024balancing,pearl2009causal} to link old and new knowledge, addressing noisy labels with curriculum learning. 
SpanKL \cite{zhang2023neural} shifts the paradigm by modeling continual named entity recognition at the \textit{span-level}, which reduces conflicts in labeling. 
SKD-NER \cite{chen2023skd} refines distillation by incorporating \textit{reinforcement learning} to optimize the selection of temperature coefficients and weights for better soft label generation.

\subsubsection{\textbf{Replay-Based}}
Although continual named entity recognition is a token-level task, the stored replay samples are at the sentence level, incorporating contextual information about the entities.

To better utilize replay samples for reviewing old entities, several works have designed different methods to extract old knowledge. 
L\&R \cite{xia2022learn} employs generative models to produce pseudo-samples that enhance the training with historical entity data. 
OCILNER \cite{ma2023learning} utilizes replay samples to calculate the centers of old-class entities and employs contrastive learning to cluster entities in the feature space, enhancing the discriminability between entities.
KD+R+K \cite{yu2021lifelong} aggregates the feature representations of new and old entities based on their similarity, initializing representations for new entities and enhancing the associations between new and old entities.
To improve \textit{storage efficiency}, KCN \cite{cao2020incremental} leverages the similarity between replay samples and class centers to gradually prune old samples that are far from the class centers while continuously adding new samples.

\subsubsection{\textbf{Prototype-Based}}
Compared to replay-based methods, prototype-based approaches often employ clustering centers or class means to define prototypes, avoiding the direct use of old samples. This approach mitigates concerns about privacy and storage limitations to some extent.
SDAPN \cite{chen2022similarity} assigns portions of the feature space to new classes preemptively and uses the similarity between new samples and old class prototypes to correct biases. 
ProtoNER \cite{kumar2023protoner} replaces traditional linear classifiers with prototypes derived from the last hidden layer's feature vectors, refining the classification process. 
IS3 \cite{qiu2024incremental} combats semantic biases by integrating prototypes with a de-biased cross-entropy loss, ensuring the model does not disproportionately favor newer over older classes.

\subsubsection{\textbf{Architecture-Based}}
Addressing the challenge of high resource costs associated with full model fine-tuning, architecture-based methods \cite{shen2019progressive,liu2023teamwork,song2023conpet} focus on modifying the model structure to support continual learning without extensive retraining. 
ICE \cite{liu2023teamwork} maintains a static model backbone, using frozen classifiers for known entities and introducing new classifiers for emerging entities during training. At inference, these classifiers are unified to ensure comprehensive entity recognition. 
ConPET \cite{song2023conpet} employs distinct Parameter Efficient Tuning (PET) modules for each task, significantly reducing tuning overhead and minimizing both overfitting and forgetting.

\begin{table}[!t]
  \centering
  \caption{Comparison between representative methods for \textbf{continual text relation extraction} and \textbf{continual machine translation}. \textbf{PEFT} represents whether utilize parameter-efficient finetuning methods for training models. \textbf{Replay, Regularization, Distillation, Architecture} refer to the common techniques summarized in Section~\ref{sec:overview_common_techniqes}. 
  }
  \resizebox{\linewidth}{!}{
    \begin{tabular}{llllp{6cm}lllllll}
    \toprule
    \multicolumn{1}{c}{\textbf{Method}} & \multicolumn{1}{c}{\textbf{Year}} & \multicolumn{1}{c}{\textbf{Publication}} & \multicolumn{1}{c}{\textbf{Backbone}} & \multicolumn{1}{c}{\textbf{Dataset}} & \multicolumn{1}{c}{\textbf{Code}} & \multicolumn{1}{c}{\textbf{PEFT}} & \multicolumn{1}{c}{\textbf{Replay}} & \multicolumn{1}{c}{\textbf{Distillation}} & \multicolumn{1}{c}{\textbf{Regularization}} & \multicolumn{1}{c}{\textbf{Architecture}} & \multicolumn{1}{c}{\textbf{Others}} \\
    \midrule
    \rowcolor[rgb]{ .788,  .788,  .788} \multicolumn{12}{c}{\textit{\textbf{Continual Relation Extraction}}} \\
    \midrule
    MLLRE \cite{obamuyide2019meta} & 2019  & RepL4NLP & Bi-LSTM & FewRel, SimpleQuestions & /     & /     & \color[RGB]{3,191,61}{\Checkmark} & /     & /     & /     & Meta Learning \\
    \midrule
    \rowcolor[rgb]{ 1,  .949,  .8} EA-EMR \cite{wang2019sentence} & 2019  & NAACL & Bi-LSTM & FewRel, SimpleQuestions & \href{https://github.com/hongwang600/Lifelong_Relation_Detection}{Link} & /     & \color[RGB]{3,191,61}{\Checkmark} & /     & /     & /     & / \\
    \midrule
    EMAR \cite{EMAR} & 2020  & ACL   & Bi-LSTM & FewRel, SimpleQuestions, TACRED & \href{https://github.com/thunlp/ContinualRE}{Link} & /     & \color[RGB]{3,191,61}{\Checkmark} & /     & /     & /     & Prototype \\
    \midrule
    \rowcolor[rgb]{ 1,  .949,  .8} CML \cite{wu2021curriculum} & 2021  & AAAI  & Bi-LSTM & FewRel, SimpleQuestions, TACRED & \href{https://github.com/wutong8023/AAAI-CML}{Link} & /     & \color[RGB]{3,191,61}{\Checkmark} & /     & /     & /     & Meta Learning \\
    \midrule
    RP-CRE \cite{RPCRE} & 2021  & ACL   & BERT  & FewRel, TACRED & \href{https://github.com/fd2014cl/RP-CRE}{Link} & /     & \color[RGB]{3,191,61}{\Checkmark} & /     & /     & /     & Prototype \\
    \midrule
    \rowcolor[rgb]{ 1,  .949,  .8} CRL \cite{CRL} & 2022  & ACL (Findings) & BERT  & FewRel, TACRED & \href{https://github.com/thuiar/CRL}{Link} & /     & \color[RGB]{3,191,61}{\Checkmark} & \color[RGB]{3,191,61}{\Checkmark} & /     & /     & Contrastive Learning, Prototype \\
    \midrule
    ERDA \cite{ERDA} & 2022  & ACL   & Bi-LSTM, BERT & FewRel, TACRED & \href{https://github.com/qcwthu/Continual_Fewshot_Relation_Learning}{Link} & /     & \color[RGB]{3,191,61}{\Checkmark} & /     & /     & /     & Contrastive Learning, Prototype \\
    \midrule
    \rowcolor[rgb]{ 1,  .949,  .8} FEA \cite{FEA} & 2022  & /     & BERT  & FewRel, TACRED & /     & /     & \color[RGB]{3,191,61}{\Checkmark} & /     & /     & /     & / \\
    \midrule
    CRECL \cite{CRECL} & 2022  & COLING & BERT  & FewRel, TACRED & \href{https://github.com/PaperDiscovery/CRECL}{Link} & /     & \color[RGB]{3,191,61}{\Checkmark} & /     & /     & /     & Contrastive Learning, Prototype \\
    \midrule
    \rowcolor[rgb]{ 1,  .949,  .8} ACA \cite{ACA} & 2022  & EMNLP & BERT  & FewRel, TACRED & \href{https://github.com/Wangpeiyi9979/ACA}{Link} & /     & \color[RGB]{3,191,61}{\Checkmark} & /     & /     & /     & Data Augmentation \\
    \midrule
    KIP-Framework \cite{KIP} & 2022  & TASLP & BERT  & FewRel, SimpleQuestions, TACRED & /     & /     & \color[RGB]{3,191,61}{\Checkmark} & /     & /     & /     & Prototype \\
    \midrule
    \rowcolor[rgb]{ 1,  .949,  .8} ConPL \cite{chen2023consistent} & 2023  & ACL   & BERT  & FewRel, TACRED & \href{https://github.com/XiudiChen/ConPL}{Link} & Prompt Tuning & \color[RGB]{3,191,61}{\Checkmark} & /     & /     & /     & Prototype \\
    \midrule
    Xia et. al \cite{xia2023enhancing} & 2023  & ACL (Findings) & BERT  & FewRel, TACRED & \href{https://github.com/hemingkx/CDec}{Link} & /     & \color[RGB]{3,191,61}{\Checkmark} & /     & /     & /     & Adversarial Tuning \\
    \midrule
    \rowcolor[rgb]{ 1,  .949,  .8} CEAR \cite{CEAR} & 2023  & ACL   & BERT  & FewRel, TACRED & \href{https://github.com/nju-websoft/CEAR}{Link} & /     & \color[RGB]{3,191,61}{\Checkmark} & \color[RGB]{3,191,61}{\Checkmark} & /     & /     & Contrastive Learning, Prototype \\
    \midrule
    SCKD \cite{SCKD} & 2023  & ACL (Findings) & BERT  & FewRel, TACRED & \href{https://github.com/nju-websoft/SCKD}{Link} & /     & \color[RGB]{3,191,61}{\Checkmark} & \color[RGB]{3,191,61}{\Checkmark} & \color[RGB]{3,191,61}{\Checkmark} & /     & Data Augmentation \\
    \midrule
    \rowcolor[rgb]{ 1,  .949,  .8} ICE \cite{liu2023teamwork} & 2023  & ACL (Findings) & BERT  & TACRED & \href{https://github.com/VT-NLP/ICE}{Link} & /     & /     & /     & /     & \color[RGB]{3,191,61}{\Checkmark} & Frozen Backbones \\
    \midrule
    ICA-Proto \cite{jiang2023ica} & 2023  & EACL (Findings) & BERT, Glove & FewRel & /     & /     & /     & /     & /     & /     & Prototype \\
    \midrule
    \rowcolor[rgb]{ 1,  .949,  .8} SEQ* \cite{zheng2023learn} & 2023  & /     & BERT, GPT2, Pythia & FewRel, TACRED & \href{https://github.com/zzz47zzz/pretrained-lm-for-incremental-learning}{Link} & /     & /     & /     & /     & /     & Classifier Expasion \\
    \midrule
    \rowcolor[rgb]{ .788,  .788,  .788} \multicolumn{12}{c}{\textit{\textbf{Continual Machine Translation}}} \\
    \midrule
    Khayrallah et. al. \cite{khayrallah2018regularized} & 2018  & NGT   & Bi-LSTM & WMT, TED-Talks, EMEA & \href{https://github.com/khayrallah/OpenNMT-py-reg}{Link} &       & /     & /     & \color[RGB]{3,191,61}{\Checkmark} & /     & / \\
    \midrule
    \rowcolor[rgb]{ .886,  .937,  .855} Escolano et. al. \cite{escolano2019bilingual} & 2019  & JASIST & Transformer & WMT   & /     & /     & /     & /     & /     & \color[RGB]{3,191,61}{\Checkmark} & Decomposed Vector Quantization \\
    \midrule
    Barrault et. al. \cite{barrault2020findings} & 2020  & WMT   & GRU   & WMT   & /     & /     & /     & /     & /     & \color[RGB]{3,191,61}{\Checkmark} & / \\
    \midrule
    \rowcolor[rgb]{ .886,  .937,  .855} Berard et. al. \cite{berard2021continual} & 2021  & WMT   & BERT  & TED-Talks & /     & /     & /     & /     & /     & \color[RGB]{3,191,61}{\Checkmark} & Vocabulary \\
    \midrule
    Cao et. al. \cite{cao2021continual} & 2021  & NAACL & Transformer & WMT, IWSLT2013 & \href{https://github.com/caoy1996/CLNMT}{Link} & /     & \color[RGB]{3,191,61}{\Checkmark} & \color[RGB]{3,191,61}{\Checkmark} & /     & /     & / \\
    \midrule
    \rowcolor[rgb]{ .886,  .937,  .855} Garcia et. al. \cite{garcia2021towards} & 2021  & NAACL & Transformer & WMT, Paracrawl & /     & /     & /     & /     & /     & \color[RGB]{3,191,61}{\Checkmark} & Vocabulary Substitution \\
    \midrule
    COKD \cite{shao2022overcoming} & 2022  & ACL   & Transformer & WMT, IWSLT15, TED bilingual & \href{https://github.com/ictnlp/COKD}{Link} & /     & /     & \color[RGB]{3,191,61}{\Checkmark} & /     & /     & / \\
    \midrule
    \rowcolor[rgb]{ .886,  .937,  .855} COMETA \cite{zhang2022clle} & 2022  & EMNLP (Findings) & Transformer & CN-25 & \href{https://github.com/HITSZ-HLT/CLLE}{Link} & /     & /     & /     & \color[RGB]{3,191,61}{\Checkmark} & /     & Meta Learning \\
    \midrule
    LFR \cite{gu2022continual} & 2022  & EMNLP & mBART50-nn & FLORES-101, OPUS100 & \href{https://github.com/ictnlp/LFR-NMT}{Link} & /     & /     & \color[RGB]{3,191,61}{\Checkmark} & \color[RGB]{3,191,61}{\Checkmark} & /     & / \\
    \midrule
    \rowcolor[rgb]{ .886,  .937,  .855} EVS \cite{huang2022entropy} & 2022  & EMNLP & Transformer & WMT   & \href{https://github.com/koukaiu/evs}{Link} & /     & /     & /     & /     & \color[RGB]{3,191,61}{\Checkmark} & Vocabulary Substitution \\
    \midrule
    CKD \cite{zhang2023continualknowledge} & 2023  & ACL   & Transformer & LDC, AI Challenger 2018, translation2019zh, TED transcripts, Subtitles & \href{https://github.com/THUNLP-MT/CKD}{Link} & /     & /     & \color[RGB]{3,191,61}{\Checkmark} & /     & /     & / \\
    \midrule
    \rowcolor[rgb]{ .886,  .937,  .855} KT \cite{huang2023knowledge} & 2023  & ACL   & Transformer & WMT   & \href{https://github.com/THUNLP-MT/ktnmt}{Link} & /     & /     & /     & /     & \color[RGB]{3,191,61}{\Checkmark} & / \\
    \midrule
    BVP \cite{liu2023continual} & 2023  & EMNLP & mBART50-nn & WMT   & \href{https://github.com/raburabu91/BVP4CL}{Link} & /     & /     & /     & /     & \color[RGB]{3,191,61}{\Checkmark} & Pruning \\
    \midrule
    \rowcolor[rgb]{ .886,  .937,  .855} SG-Rep \cite{Resta2024selfgenerated} & 2024  & /     & T5    & IWSLT17, UNPC & \href{https://github.com/m-resta/sg-rep}{Link} & /     & Pseudo Sample & /     & /     & /     & / \\
    \midrule
    F-MALLOC \cite{wu2024f} & 2024  & NAACL & Transformer & WMT   & \href{https://github.com/WJMacro/ContinualMT.}{Link} & /     & /     & /     & /     & \color[RGB]{3,191,61}{\Checkmark} & Pruning \\
    \bottomrule
\end{tabular}%
    }
  \label{tab:RE_MT_methods}%
\end{table}%
\subsection{Continual Relation Extraction}
\label{sec:continual_relation_extraction}

Continual Relation Extraction (CRE) entails updating relation extraction models to recognize new relationships while retaining accuracy on previously learned data. For instance, from the sentence "Lange was born July 23, 1957, in Illinois," a relation extraction system identifies the relationship between "Lange" and "Illinois" as "State or Province of Birth." The challenge is for the system to learn new relationships, like "Country of Headquarters," without forgetting existing ones. Apart from catastrophic forgetting, continual relation extraction confronts two challenges: (1) Order Sensitivity \cite{yoon2019scalable,chen2018lifelong}: This refers to the phenomenon where the performance of the model varies depending on the sequence of task introduction. (2) Interference of Analogous Relations \cite{ACA,CEAR}: Challenges arise when the model confuses similar relations, such as "country of headquarters" and "state or province of headquarters."

In continual relation extraction, experience replay \cite{EMAR,RPCRE,CRL,CRECL} are widely favored due to their efficacy in managing both the acquisition of new information and the retention of old knowledge. Five popular techniques are combined with experience replay: \emph{knowledge distillation} \cite{CEAR,SCKD,CRL}, \emph{relation prototypes} \cite{EMAR,RPCRE,FEA,KIP}, \emph{contrastive learning} \cite{CRECL,CRL,CPL,CEAR}, \emph{meta learning} \cite{obamuyide2019meta,wu2021curriculum}, \emph{data augmentation} \cite{ACA,ERDA,CPL}. Table~\ref{tab:RE_MT_methods} provides a detailed comparison of these methods.

\subsubsection{\textbf{Knowledge Distillation}}
Focal Knowledge Distillation (FKD) is utilized by CEAR\cite{CEAR}. Specifically, FKD focuses on assigning higher importance to analogous relations, whereas SCKD \cite{SCKD} emphasizes serial distillation with pseudo-samples to bolster few-shot learning capabilities. 
In contrast, the focus on consistent relation representation learning across tasks makes CRL \cite{CRL} align embedding in memory maintenance and ensure stability in the embedding space.

\subsubsection{\textbf{Relation Prototypes}}
Relation prototypes refer to a representation of relation in the feature space. As a pioneer work, EMAR \cite{EMAR} focus on utilizing relation prototypes for memory replay. 
Similarly, relation prototypes are used by RP-CRE \cite{RPCRE} to refine sample embeddings. 
Inspired by EMAR \cite{EMAR} and RP-CRE \cite{RPCRE}, a more simplified variant in FEA \cite{FEA} operates through the fast adaption and balanced tuning process. 
With the help of external knowledge, KIP-Framework\cite{KIP} infuses prototypes with these knowledge to generate prototypes.

\subsubsection{\textbf{Contrastive Learning}}
The application of contrastive learning \cite{jaiswal2020survey} varies from focusing on data distribution and embedding stability (CRECL \cite{CRECL} and CRL \cite{CRL}) to addressing few-shot learning and overfitting challenges (CPL\cite{CPL}), as well as enhancing the distinction of analogous relations (CEAR\cite{CEAR}).
CRECL\cite{CRECL} uses a contrastive network, which contrasts a given instance with prototypes of each candidate relation stored in a memory module. For contrastive replay, it is used by CRL\cite{CRL} to train memorized samples. Similarly, CEAR\cite{CEAR} utilizes contrastive learning alongside a linear method for training, where the former helps in improving feature space alignment and the latter ensures task-specific decision boundaries. Besides, a margin-based contrastive learning objective is introduced by CPL \cite{CPL} to gain discriminative representations.

\subsubsection{\textbf{Meta Learning}}
To enable models to adapt quickly to new tasks while mitigating catastrophic forgetting, MLLRE \cite{obamuyide2019meta} and CML \cite{wu2021curriculum} both use meta-learning frameworks. 
On the one hand, MLLRE\cite{obamuyide2019meta} employs the REPTILE algorithm \cite{nichol2018first} for gradient-based meta-learning without second-order derivatives. 
On the other hand, CML\cite{wu2021curriculum} combines curriculum learning with meta learning to create a dynamic learning curriculum that prioritizes tasks based on difficulty. The main difference is that CML\cite{wu2021curriculum} focuses on task ordering and the difficulty in constructing learning curricula, while MLLRE\cite{obamuyide2019meta} directly optimizes meta-objectives.

\subsubsection{\textbf{Data Augmentation}}
Data augmentation is leveraged to enrich the training data and improve model generalization across tasks, especially in low-resource settings. The majority of methods utilize external data\cite{ERDA} or generated samples\cite{ACA,CPL}.
Adversarial examples are incorporated by ACA\cite{ACA} to enhance model robustness and generalization.
Besides, ERDA\cite{ERDA} selects informative samples from an unlabeled corpus that consists of sentences from Wikipedia to provide more relational knowledge for few-shot tasks.
With the help of large language models, CPL \cite{CPL} guides them to generate diverse and relevant samples for memory augmentation.

\subsection{Continual Machine Translation}
\label{sec:continual_machine_translation}

Continual Machine Translation \cite{cao2021continual, garcia2021towards,castellucci2021learning, huang2022entropy,gu2022continual,shao2022overcoming,winata2023overcoming,huang2023knowledge,zhang2023towards} is devised to cater to the demands of multilingual tasks in real-world scenarios, facilitating the addition of new languages over time. Continual machine translation typically undergoes training on a general domain corpus, encompassing a collection of various languages, followed by fine-tuning through continued training on an in-domain corpus specific to new languages. The objective is to learn the new language while retaining knowledge of the initial languages. Most of the methods for continual machine translation are single-step incremental language learning \cite{cao2021continual, garcia2021towards,castellucci2021learning, huang2022entropy,gu2022continual,shao2022overcoming,huang2023knowledge,zhang2023towards}, and a small number are multi-step incremental language learning \cite{castellucci2021learning,winata2023overcoming}. Several articles contribute to the field by proposing new benchmarks tailored to assess lifelong learning capabilities in multilingual contexts. Barrault et al. \cite{barrault2020findings} provides training, lifelong, and test datasets for English-German and English-French to push forward research in lifelong learning NMT. Conversely, CLLE \cite{zhang2022clle} introduces a Chinese-centric benchmark, featuring tasks that test a model’s ability to handle closely related languages and diverse language families, reflecting real-world demands. Furthermore, Continual machine translation methods can be broadly classified into four primary approaches: \emph{distillation-based} \cite{cao2021continual,shao2022overcoming,zhang2023continualknowledge}, \emph{regularization-based} \cite{khayrallah2018regularized,gu2022continual,liu2023continual}, \emph{architecture-based} \cite{garcia2021towards,berard2021continual,huang2022entropy,huang2023knowledge,wu2024f}, and others \cite{barrault2020findings,zhang2022clle,Resta2024selfgenerated,escolano2019bilingual}. A detailed comparison between these methods is provided in Table~\ref{tab:RE_MT_methods}.

\subsubsection{\textbf{Distillation-Based}}
Traditional NMT models are unable to handle continual or sequential learning problems without forgetting previously learned knowledge. Therefore, there are several methods innovating with different facets of dynamic knowledge distillation, such as Cao et al. \cite{cao2021continual} and CKD \cite{zhang2023continualknowledge}. In addition to this, to address the unbalanced training problem, COKD \cite{shao2022overcoming} balances the model's focus across training samples, uniquely integrating dynamically updated teacher models.

\subsubsection{\textbf{Regularization-Based}}
To balance learning objectives on continual neural machine translation, there are many different implementations, such as regularizing the training process to minimize deviation from established models \cite{khayrallah2018regularized}, identifying parameter updates that risk minimal forgetting \cite{gu2022continual}, or categorizing parameters based on their relevance to specific tasks or overall functionality \cite{liu2023continual}.

\subsubsection{\textbf{Architecture-Based}}
Architecture-based approaches in machine translation include \textit{lexical structure} \cite{garcia2021towards,berard2021continual,huang2022entropy} and \textit{model structure} \cite{escolano2019bilingual, huang2023knowledge,wu2024f}. 

\textit{Lexical structure} refers to the set of unique tokens or words that an NMT model can recognize and generate. These tokens typically include words, subwords, or characters that the model uses to process and translate text from one language to another. EVS \cite{huang2022entropy} optimizes embedding spaces by dynamically managing vocabularies based on their entropy values across languages, enhancing linguistic diversity without enlarging the model. Similarly, the method proposed by Garcia et al. \cite{garcia2021towards} refines embedding efficiency by selectively substituting vocabulary parts, maintaining translation quality while integrating new languages efficiently.

\textit{Model structure} innovations are highlighted by the introduction of dynamic resource allocation mechanisms and modular adaptation, which determines how effectively a model can handle different linguistic elements, especially when translating between multiple languages. F-MALLOC \cite{wu2024f} introduces a memory allocation model that adapts to new languages by dynamically adjusting resources, thus supporting scalable and efficient learning. Concurrently, KT \cite{huang2023knowledge} integrates language-specific adapters into the NMT framework, facilitating seamless knowledge transfer and enabling the model to learn new languages without extensive retraining, thereby preserving its performance across a diverse linguistic spectrum.
\begin{table}[!t]
  \centering
  \caption{Comparison between representative methods for \textbf{continual instruction tuning}, \textbf{continual knowledge editing}, and \textbf{continual alignment}. \textbf{PEFT} represents whether utilize parameter-efficient finetuning methods for training models. \textbf{Replay, Regularization, Distillation, Architecture} refer to the common techniques summarized in Section~\ref{sec:overview_common_techniqes}. 
  }
  \resizebox{\linewidth}{!}{
    \begin{tabular}{llllp{6cm}lllllll}
    \toprule
    \multicolumn{1}{c}{\textbf{Method}} & \multicolumn{1}{c}{\textbf{Year}} & \multicolumn{1}{c}{\textbf{Publication}} & \multicolumn{1}{c}{\textbf{Backbone}} & \multicolumn{1}{c}{\textbf{Dataset}} & \multicolumn{1}{c}{\textbf{Code}} & \multicolumn{1}{c}{\textbf{PEFT}} & \multicolumn{1}{c}{\textbf{Replay}} & \multicolumn{1}{c}{\textbf{Distillation}} & \multicolumn{1}{c}{\textbf{Regularization}} & \multicolumn{1}{c}{\textbf{Architecture}} & \multicolumn{1}{c}{\textbf{Others}} \\
    \midrule
    \rowcolor[rgb]{ .788,  .788,  .788} \multicolumn{12}{c}{\textit{\textbf{Continual Instruction Tuning}}} \\
    \midrule
    IDS \cite{wang2019incremental} & 2019  & ACL   & GRU   & SubD1-D5 & \href{https://github.com/Leechikara/Incremental-Dialogue-System}{Link} & /     & /     & /     & /     & /     & Uncertainty Estimation \\
    \midrule
    \rowcolor[rgb]{ .937,  .933,  .992} DnR \cite{sun2020distill} & 2020  & COLING & GPT-2 & SST, QA-SRL, WOZ, SQUAD, WIkiSQL, AGNews, Yelp, Amazon, DBPedia, Yahoo & /     & /     & Pseudo Sample & \color[RGB]{3,191,61}{\Checkmark} & /     & /     & / \\
    \midrule
    ARPER \cite{mi2020continual} & 2020  & EMNLP (Findings) & GPT-2 & MultiWoZ-2.0 & \href{https://github.com/MiFei/Continual-Learning-for-NLG}{Link} & /     & \color[RGB]{3,191,61}{\Checkmark} & /     & \color[RGB]{3,191,61}{\Checkmark} & /     & / \\
    \midrule
    \rowcolor[rgb]{ .937,  .933,  .992} LAMOL \cite{sun2019lamol} & 2020  & ICLR  & GPT-2 & SST, QA-SRL, WOZ, SQUAD, WIkiSQL, AGNews, Yelp, Amazon, DBPedia, Yahoo & \href{https://github.com/jojotenya/LAMOL}{Link} & /     & Pseudo Sample & /     & /     & /     & / \\
    \midrule
    Rational LAMOL \cite{kanwatchara2021rational} & 2021  & ACL   & GPT-2 & BoolQ, Movie, SciFact & \href{https://github.com/kanwatchara-k/r_lamol}{Link} & /     & Pseudo Sample & /     & /     & /     & / \\
    \midrule
    \rowcolor[rgb]{ .937,  .933,  .992} TPEM \cite{geng2021continual} & 2021  & ACL   & GRU   & In-Car Assistant, Multi-WOZ 2.1, CamRest & \href{https://github.com/siat-nlp/TPEM}{Link} & /     & /     & /     & /     & \color[RGB]{3,191,61}{\Checkmark} & Pruning \\
    \midrule
    BiHNet \cite{jin2021learn} & 2021  & EMNLP (Findings) & BART  & CLIF-26, CLIF-55 & \href{https://github.com/INK-USC/CLIF}{Link} & Adapters & /     & /     & \color[RGB]{3,191,61}{\Checkmark} & /     & Hyper-Networks \\
    \midrule
    \rowcolor[rgb]{ .937,  .933,  .992} AdapterCL \cite{madotto2021continual} & 2021  & EMNLP & GPT-2 & TaskMaster 2019, TaskMaster 2020, Schema\newline{}Guided Dialogue, MultiWoZ & \href{https://github.com/andreamad8/ToDCL}{Link} & Adapters & /     & /     & /     & \color[RGB]{3,191,61}{\Checkmark} & / \\
    \midrule
    ACM \cite{zhang2022continual} & 2022  & ACL   & GPT-2 & E2ENLG, RNNLG, WikiSQL, CNN/DailyMail, MultiWOZ & \href{https://github.com/GT-SALT/Adaptive-Compositional-Modules}{Link} & Adapters & Pseudo Sample & /     & /     & \color[RGB]{3,191,61}{\Checkmark} & / \\
    \midrule
    \rowcolor[rgb]{ .937,  .933,  .992} InstructionSpeak \cite{yin2022contintin} & 2022  & ACL   & BART  & NaturalInstructions & /     & /     & \color[RGB]{3,191,61}{\Checkmark} & /     & /     & /     & / \\
    \midrule
    Continual Prompt Tuning \cite{zhu2022continual} & 2022  & ACL   & T5    & Schema Guided Dialogue & \href{https://github.com/thu-coai/cpt4dst}{Link} & Prompt Tuning & \color[RGB]{3,191,61}{\Checkmark} & /     & /     & /     & / \\
    \midrule
    \rowcolor[rgb]{ .937,  .933,  .992} PCLL \cite{zhao2022prompt} & 2022  & EMNLP & GPT-2 & DSTC, TOP & \href{https://github.com/AlibabaResearch/DAMO-ConvAI/tree/main/pcll}{Link} & /     & Pseudo Sample & \color[RGB]{3,191,61}{\Checkmark} & /     & /     & Variational Auto Encoder \\
    \midrule
    CT0 \cite{scialom2022fine} & 2022  & EMNLP & T0    & Simpl, HGen, Haiku, CQA, InqQG, EmDg, Exp, TwSt & \href{https://github.com/ThomasScialom/T0_continual_learning}{Link} & /     & \color[RGB]{3,191,61}{\Checkmark} & /     & /     & /     & / \\
    \midrule
    \rowcolor[rgb]{ .937,  .933,  .992} LFPT5 \cite{qin2021lfpt5} & 2022  & ICLR  & T5    & AGNews, Amazon Review, DBPedia, Yahoo, CNNDM, WikiHow, Xsum & \href{https://github.com/qcwthu/Lifelong-Fewshot-Language-Learning}{Link} & Prompt Tuning & Pseudo Sample & /     & /     & /     & / \\
    \midrule
    LPT \cite{liang2023prompts} & 2023  & ACL   & T5    & ACE05-Ent, CoNLL03, CoNLL04, ACE05Rel, SciERC,NYT, CASIE, ACE05-Evt, SemEval-14, SemEval-15, SemEval-16 & \href{https://github.com/jokieleung/Lottery_Prompt}{Link} & Prompt Tuning & /     & /     & /     & \color[RGB]{3,191,61}{\Checkmark} & Pruning \\
    \midrule
    \rowcolor[rgb]{ .937,  .933,  .992} DYNAINST \cite{mok2023large} & 2023  & ACL   & BART  & SuperNI & /     & /     & \color[RGB]{3,191,61}{\Checkmark} & /     & /     & /     & / \\
    \midrule
    HMI-LAMOL \cite{maekawa2023generative} & 2023  & EACL  & GPT-2, BERT & SQuAD, WikiSQL, SST, QASRL, WOZ, AGNews, Yelp, Amazon, DBPedia, Yahoo & \href{https://github.com/arumaekawa/GR-HMI}{Link} & /     & Pseudo Sample & /     & /     & /     & / \\
    \midrule
    \rowcolor[rgb]{ .937,  .933,  .992} DMEA \cite{qin2023lifelong} & 2023  & EMNLP & GPT-2, BERT & RNNLG, E2ENLG, CNN/DailyMail, MultiWOZ, WikiSQL & /     & Adapters & /     & /     & /     & /     & / \\
    \midrule
    O-LoRA \cite{wang2023orthogonal} & 2023  & EMNLP (Findings) & LLaMA, T5 & GLUE, SuperGLUE, IMDB & \href{https://github.com/cmnfriend/O-LoRA}{Link} & LoRA  & /     & /     & \color[RGB]{3,191,61}{\Checkmark} & \color[RGB]{3,191,61}{\Checkmark} & Orthogonal Subspaces \\
    \midrule
    \rowcolor[rgb]{ .937,  .933,  .992} TSS \cite{ke2023sub} & 2023  & EMNLP (Findings) & BART  & AGNews, Yelp, Amazon, DBPedia, Yahoo & \href{https://github.com/ZixuanKe/PyContinual}{Link} & Adapters & /     & /     & /     & \color[RGB]{3,191,61}{\Checkmark} & / \\
    \midrule
    ProgPrompt \cite{razdaibiedina2022progressive} & 2023  & ICLR  & T5, BERT & GLUE, SuperGLUE, IMDB & \href{https://github.com/arazd/ProgressivePrompts}{Link} & Prompt Tuning & /     & /     & /     & \color[RGB]{3,191,61}{\Checkmark} & / \\
    \midrule
    \rowcolor[rgb]{ .937,  .933,  .992} SAPT \cite{modulesapt} & 2024  & /     & LLaMA, T5 & SuperNI, GLUE, SuperGLUE, IMDB & /     & Prompt Tuning, LoRA & Pseudo Sample & /     & /     & \color[RGB]{3,191,61}{\Checkmark} & / \\
    \midrule
    InsCL \cite{wang2024inscl} & 2024  & /     & LLaMA & SuperNI & /     & /     & \color[RGB]{3,191,61}{\Checkmark} & /     & /     & /     & / \\
    \midrule
    \rowcolor[rgb]{ .937,  .933,  .992} I-LoRA \cite{ren2024analyzing} & 2024  & /     & LLaMA & ScienseQA, MedMCQA, FOMC, JEC-QA, C-STANCE, 20Minuten, NumGLUE, MMLU, BBH, PIQA & \href{https://github.com/which47/LLMCL}{Link} & LoRA  & \color[RGB]{3,191,61}{\Checkmark} & \color[RGB]{3,191,61}{\Checkmark} & /     & \color[RGB]{3,191,61}{\Checkmark} & / \\
    \midrule
    SSR \cite{huang2024mitigating} & 2024  & /     & LLaMA, Alpaca & SuperNI & /     & LoRA  & Pseudo Sample & /     & /     & /     & / \\
    \midrule
    \rowcolor[rgb]{ .937,  .933,  .992} SLM \cite{bohaoscalable} & 2024  & ICLR  & LLaMA, T5, BERT & AGNews, Yelp, Amazon, DBPedia, Yahoo, Medical, MMLU, Finance & \href{https://github.com/Pbihao/SLM}{Link} & LoRA  & /     & /     & /     & /     & / \\
    \midrule
    Q-Tuning \cite{guo2024q} & 2024  & NAACL (Findings) & BERT, T5 & GLUE, SuperGLUE, IMDB & /     & Prompt Tuning & /     & /     & /     & \color[RGB]{3,191,61}{\Checkmark} & / \\
    \midrule
    \rowcolor[rgb]{ .937,  .933,  .992} SAPT \cite{modulesapt} & 2024  & /     & T5, LLaMA & SuperNI, GLUE, SuperGLUE, IMDB & /     & LoRA, Prompt Tuning & Pseudo Sample & \color[RGB]{3,191,61}{\Checkmark} & \color[RGB]{3,191,61}{\Checkmark} & /     & / \\
    \midrule
    MoRAL \cite{moral} & 2024  & /     & LLaMA, Phi & Arxiv, HotpotQA & /     & LoRA  & /     & /     & /     & \color[RGB]{3,191,61}{\Checkmark} & / \\
    \midrule
    \rowcolor[rgb]{ .788,  .788,  .788} \multicolumn{12}{c}{\textit{\textbf{Continual Knowledge Editing}}} \\
    \midrule
    Lee et. al. \cite{lee2022plug} & 2022  & ACL (Findings) & T5    & zsRE, NQ-SituatedQA & \href{https://github.com/wookjeHan/Continual-Plug-and-Adapt-for-CuQA/}{Link} & LoRA, K-Adapter & /     & /     & \color[RGB]{3,191,61}{\Checkmark} & \color[RGB]{3,191,61}{\Checkmark} & / \\
    \midrule
    \rowcolor[rgb]{ .992,  .937,  .878} SLAG \cite{SLAG} & 2023  & EACL  & BART, RoBERTa & zsRE, Wikidata5m, FEVER, LeapOfThought & \href{https://github.com/peterbhase/SLAG-Belief-Updating}{Link} & /     & /     & /     & /     & /     & / \\
    \midrule
    GRACE \cite{GRACE} & 2023  & ICLR  & T5, BERT & zsRE, SCOTUS, Natural Questions & /     & GRACE Adapters & /     & /     & /     & \color[RGB]{3,191,61}{\Checkmark} & Codebook \\
    \midrule
    \rowcolor[rgb]{ .992,  .937,  .878} TPatcher \cite{TPatcher} & 2023  & ICLR  & BART, BERT & zsRE, FEVER, CBQA & \href{https://github.com/ZeroYuHuang/Transformer-Patcher}{Link} & /     & /     & /     & /     & \color[RGB]{3,191,61}{\Checkmark} & / \\
    \midrule
    WilKE \cite{WilKE} & 2024  & /     & GPT-J, GPT-2 & CounterFact & /     & /     & /     & /     & /     & \color[RGB]{3,191,61}{\Checkmark} & / \\
    \midrule
    \rowcolor[rgb]{ .788,  .788,  .788} \multicolumn{12}{c}{\textit{\textbf{Continual Alignment}}} \\
    \midrule
    Zhao et. al. \cite{zhao2023learning} & 2023  & /     & LLaMA, GPT-2 & BBQ, Pile, HarmfulQA & /     & LoRA  & \color[RGB]{3,191,61}{\Checkmark} & /     & /     & /     & Data Filtering, Self-Correction \\
    \midrule
    \rowcolor[rgb]{ .839,  .863,  .894} CPPO \cite{zhangcppo} & 2024  & ICLR  & LLaMA, GPT-2 & HH-RLHF, Reddit TL;DR & \href{https://openi.pcl.ac.cn/Hanlard/CPPO}{Link} & /     & /     & \color[RGB]{3,191,61}{\Checkmark} & /     & /     & / \\
    \midrule
    COPR \cite{zhang2024copr} & 2024  & /     & LLaMA, GPT-J, OPT, & HH-RLHF, Reddit TL;DR, IMDB & \href{https://openi.pcl.ac.cn/Hanlard/Offline_alignment_methods_based_on_trlx.git}{Link} & /     & /     & /     & \color[RGB]{3,191,61}{\Checkmark} & /     & / \\
    \bottomrule
\end{tabular}%
    }
  \label{tab:task_agnostic_methods}%
\end{table}%
\subsection{Continual Instruction Tuning}
\label{sec:continual_instruction_tuning}

The traditional machine learning paradigm for NLP assumed that target tasks were predefined and static, and that task supervision relied on labeled samples. This raises the question of how to build a system that can continuously learn new tasks from their instructions. Continual Instruction Tuning addresses this by designing various instructions for the same model to solve multiple NLP tasks. Earlier literature using GPT-2 \cite{radford2018improving} often used simple instructions like dataset names or special tokens \cite{sun2019lamol}. In this survey, instruction tuning is defined more broadly, encompassing methods evaluated on a variety of generation tasks.

Chen et al. \cite{chen2024coin} proposes a comprehensive benchmark test, the Continuous Instruction tuNing (CoIN), to evaluate existing models in the sequential instruction tuning paradigm. CoIN evaluates two aspects: instruction following and general knowledge. It consists of 10 commonly used datasets spanning 8 task categories, ensuring a diverse range of instructions and tasks. Continual instruction tuning methods can be broadly classified into three primary approaches: \emph{replay-based} \cite{sun2019lamol, kanwatchara2021rational,sun2020distill,qin2021lfpt5,zhao2022prompt,maekawa2023generative,huang2024mitigating}, \emph{regularization-based} \cite{jin2021learn,mi2020continual,wang2024inscl,wang2023orthogonal,bohaoscalable}, \emph{gradient-based} \cite{lee2021sequential,korbak2022controlling}, and \emph{architecture-based} \cite{geng2021continual,wang2019incremental,zhang2022continual,madotto2021continual,ke2023sub,liang2023prompts,razdaibiedina2022progressive,zhu2022continual,scialom2022fine,yin2022contintin,guo2024q}. A detailed comparison between these methods is provided in Table~\ref{tab:task_agnostic_methods}.

\subsubsection{\textbf{Replay-Based}}
The Replay-Based methods include the Generative Replay-Based method \cite{sun2019lamol, kanwatchara2021rational,sun2020distill,qin2021lfpt5,zhao2022prompt,maekawa2023generative,huang2024mitigating} and the Experience Replay-Based method \cite{ren2024analyzing}.

Generative replay inspired by hippocampal memory mechanisms \cite{shin2017continual}, this foundational paper introduces a novel approach by mimicking the human hippocampus, renowned for its role in memory formation and recall. The model efficiently retains prior knowledge while assimilating new information, setting a baseline for addressing catastrophic forgetting. Progressing from this foundation, LAMOL \cite{sun2019lamol} embeds the generative replay directly within the language model. This integration simplifies the architecture and enables dynamic pseudo-sample generation, enhancing memory consolidation without extra computational overhead. Further refining this approach, LFPT5 \cite{qin2021lfpt5} utilizes prompt tuning to adapt quickly to new tasks with few examples, significantly reducing the data dependency and maintaining performance across tasks. Futhermore, there are several methods to improve the framework of Generative replay, such as PCLL \cite{zhao2022prompt}, HMI-LAMOL \cite{maekawa2023generative}, SSR \cite{huang2024mitigating}. A few approaches follow the conventional setting using experience replay, such as I-LoRA \cite{ren2024analyzing}.

\subsubsection{\textbf{Regularization-Based}}
The regularization-based methods can be can be broadly categorized into direct \cite{mi2020continual, wang2023orthogonal} and indirect \cite{jin2021learn,bohaoscalable,wang2024inscl} regularization approaches.

Direct regularization directly influencing model parameters to preserve prior learning. For instance, ARPER \cite{mi2020continual} integrates adaptive regularization directly into the training process, utilizing regularization terms that directly mitigate the forgetting of previously acquired knowledge during the learning of new dialogue tasks. Similarly, O-LoRA \cite{wang2023orthogonal} employs an orthogonal low-rank adaptation (O-LoRA) method that directly constrains gradient updates to be orthogonal to the subspaces of previous tasks.

Indirect regularization utilizes factors such as similarity and importance between tasks to impose indirect restrictions on model parameters. For example, BiHNet \cite{jin2021learn} leverages a bi-level hypernetwork to create task-specific adapters, an architectural adjustment that indirectly preserves past knowledge by minimizing task interference. InsCL \cite{wang2024inscl} utilizes dynamic replay of enriched data, indirectly facilitating continual learning by reintroducing crucial features of past tasks. Additionally, SLM \cite{bohaoscalable} introduces a dynamic re-parameterization mechanism that adjusts the model's parameters according to the task distribution, ensuring that each task's learning is compartmentalized, thereby reducing the overwrite of important historical information. 

\subsubsection{\textbf{Gradient-Based}} 
In the realm of continual instruction tuning, effectively managing knowledge transfer and mitigating catastrophic forgetting are critical challenges that influence the robustness and versatility of language models. Some advances have focused on innovative gradient manipulation techniques to address these issues. Lee et al. \cite{lee2021sequential} proposes a method that enhances gradient alignment across different tasks to promote better generalization and minimize negative transfer. 
Complementarily, Korbak et al. \cite{korbak2022controlling} introduces a framework for dynamically adjusting the learning parameters to preserve previously acquired knowledge during the fine-tuning process. Together, these methodologies underscore the potential of sophisticated gradient strategies to refine the adaptability of language models across diverse linguistic tasks without compromising their performance on previously learned information. 

\subsubsection{\textbf{Architecture-Based}}
Architecture-based approaches can be categorized into \textit{model-based} \cite{geng2021continual,wang2019incremental}, \textit{adapter-based} \cite{zhang2022continual,madotto2021continual,qin2023lifelong,ke2023sub,modulesapt} and \textit{prompt-based} methods\cite{liang2023prompts,razdaibiedina2022progressive,zhu2022continual,scialom2022fine,yin2022contintin,guo2024q,ren2024analyzing}.

\textit{Model-based} methods dynamically adjust the full network architecture in response to new information without requiring complete system retraining. For instance, TPEM \cite{geng2021continual} employs a cycle of pruning to eliminate less useful connections, expanding the network to accommodate new tasks, and masking to selectively deactivate certain pathways, ensuring that the system remains efficient and relevant to current tasks. Besides, Wang et al. \cite{wang2019incremental} leverages an uncertainty estimation to decide when the system should update itself and an online learning component that facilitates immediate integration of new data into the model.

\textit{Adapter-based} methods selectively adds new modules to manage knowledge retention and adaptability across sequential tasks. Several approaches allows the model to expand by dynamically adjusting and optimizing its architecture for each new task, such as ACM \cite{zhang2022continual}, DMEA \cite{qin2023lifelong} and so on. It incorporates new modules and adapts existing ones based on their performance and relevance to ongoing and past tasks, making the expansion process both targeted and efficient. In addition to this, SAPT \cite{modulesapt} does not expand by adding new layers or modules in a conventional sense, but rather by utilizing a flexible attention mechanism to apply different sets of parameters stored from previous tasks to new tasks.

\textit{Prompt-based} methods are essentially task-specific modifiers that guide the pre-trained language models in generating outputs that are appropriate for new tasks while retaining the capability to perform well on older tasks. This is achieved by strategically modifying or augmenting the input space of the models with prompts that encapsulate the essence of the task at hand, allowing the core model parameters to remain unchanged. For example, LPT \cite{liang2023prompts} uses a binary prompt mask to selectively prune ineffective prompt vectors, enhancing computational efficiency and preserving crucial task-specific knowledge. Complementarily, DYNAINST \cite{mok2023large} integrates a dynamic replay mechanism to selectively maintain training examples that improve learning efficiency, thereby optimizing knowledge retention across tasks. Further, ProgPrompt \cite{razdaibiedina2022progressive} innovates by sequentially concatenating task-specific prompts to accumulate knowledge and facilitate forward transfer without losing prior information. Together, these methods advance prompt-based strategies to boost the scalability and efficiency of lifelong learning in language models.

\subsection{Continual Knowledge Editing}
\label{sec:continual_knowledge_editing}

Continual Knowledge Editing serves as a pivotal component of lifelong learning for language models, designed to ensure their adaptability and accuracy as they encounter new information or discover that previous knowledge has become outdated \cite{TPatcher}. Unlike traditional question-answering tasks that respond based on fixed knowledge, continual knowledge editing involves updating the model's understanding through knowledge triplets—structured data forms like \emph{(head\_entity, relation, tail\_entity)}—which help in precisely defining the modifications needed in the model's knowledge base \cite{CKEsurvey}. For instance, consider the triplet (Pluto, IsA, Planet), which may need updating to (Pluto, IsA, Dwarf Planet) as astronomical definitions evolve. 

Research in this area has traditionally focused on one-step editing techniques \cite{de2021editing,dai2021knowledge,meng2022mass,mitchell2021fast,meng2022locating}, where models undergo \emph{a single, significant update} to rectify or enhance their knowledge bases. However, more recent approaches \cite{TPatcher,SLAG,WilKE,lee2022plug,GRACE} advocate for \emph{a continual and sequential editing process}, aligning more closely with the principles of lifelong learning. This involves making multiple, smaller adjustments over time, allowing the model to adapt to the changing real-world requirements and maintain its relevance and accuracy without the need for comprehensive retraining.

Continual knowledge editing methods can be categorized into three main strategies \cite{CKEsurvey}: \emph{External Memorization}, \emph{Global Optimization}, and \emph{Local Modification}. A detailed comparison between these methods is provided in Table~\ref{tab:task_agnostic_methods}.
(1) \textbf{External Memorization} methods like GRACE \cite{GRACE} and T-Patcher \cite{TPatcher} use extension-based strategies to integrate new data \cite{CKEsurvey}. GRACE, for example, employs key-value pairs to dynamically store new information, allowing the model to access the latest data without a full retraining cycle. T-Patcher, on the other hand, makes precise, targeted adjustments to model parameters to correct specific errors, similar to software patches fixing bugs, thus ensuring that the model's outputs remain accurate and current.
(2) \textbf{Global Optimization} involves more comprehensive updates across the model’s parameters, exemplified by SLAG \cite{SLAG}, which uses intermediate fine-tuning strategies to carefully balance the integration of new information with the retention of existing knowledge \cite{CKEsurvey}. This approach allows for gradual updates that refine the model's understanding without overwhelming the previously learned data. Lee et al. \cite{lee2022plug} further this concept by incorporating LoRA to focus on expanding specific parts of the model's architecture, minimizing the disruption to the overall system.
(3) \textbf{Local Modification} focuses on making changes at a more granular level within the model, such as adjusting specific neurons or layers that are most relevant to the new information. WilKE \cite{WilKE} utilizes gradient-based strategies to precisely identify and modify the parts of the model that directly relate to outdated or incorrect information \cite{CKEsurvey}, enabling targeted updates that do not require extensive retraining but still ensure the model's growth in knowledge and capabilities.
\subsection{Continual Alignment}
\label{sec:continual_alignment}

Continual alignment in Large Language Models is essential to ensure that these models remain aligned with human values and societal norms throughout their lifecycle. Traditionally, alignment has been a \emph{one-step} process where LLMs are aligned after pretraining and instruction tuning stages \cite{shen2023large}. However, as the demands and expectations from AI systems evolve, it is becoming increasingly necessary to adopt a \emph{multi-step} alignment approach \cite{zhang2024copr,zhangcppo,zhao2023learning}, where models are realigned periodically to accommodate new ethical standards and societal values. The \emph{alignment tax}, which refers to the trade-off between aligning models to human values and potentially compromising their general performance, is a critical consideration in this process \cite{lin2023mitigating}.

Continual alignment can be categorized into two main areas: \emph{value alignment} \cite{zhangcppo,zhang2024copr,lin2023mitigating} and \emph{security alignment} \cite{zhao2023learning,zhan2023removing,qi2023fine}. A detailed comparison between these methods is provided in Table~\ref{tab:task_agnostic_methods}.
(1) In \textbf{Value Alignment}, the focus is on ensuring that the model's responses adhere to ethical guidelines without losing previously acquired capabilities. Techniques such as CPPO \cite{zhangcppo} implement weighting strategies to balance new ethical priorities with existing knowledge. COPR \cite{zhang2024copr} addresses catastrophic forgetting in the context of value alignment by dynamically adjusting regularization based on both new and historical preferences. Meanwhile, Lin et al. \cite{lin2023mitigating} suggest model averaging to effectively manage the alignment tax, optimizing the balance between maintaining performance and adhering to updated values.
(2) \textbf{Security Alignment} concentrates on safeguarding the integrity and security of the data processed by LLMs. It involves strategies to prevent the perpetuation of harmful information and protect against data leaks. Zhao et al. \cite{zhao2023learning} have developed a forgetting filter technique that prioritizes the security of content during model updates. Zhan et al. \cite{zhan2023removing} demonstrate the ease with which minimal fine-tuning can compromise established security measures, highlighting the ongoing need for robust protection mechanisms. To strengthen LLMs against potential misuse and evolving security threats, ongoing research, and methodological innovations are crucial, as noted by Lermen et al. \cite{lermen2023lora} and Qi et al. \cite{qi2023fine}. These efforts ensure that as LLMs are aligned with new security protocols, they do not become vulnerable to novel forms of exploitation.

\subsection{Summary}
Building on continual pretraining, which enhances the internal knowledge of LLMs, continual finetuning further adapts these models to specific tasks such as text classification, named entity recognition, relation extraction, machine translation, and instruction tuning. Techniques like distillation, replay, regularization, architecture-based, and gradient-based methods are employed to address challenges like catastrophic forgetting and task interference. Despite advancements, significant challenges remain, particularly in maintaining long-term performance and resource efficiency. Future research should focus on innovative solutions to mitigate forgetting, enhance task adaptability, and develop efficient, scalable architectures for sustained performance across diverse tasks.

\section{Methodology: External Knowledge}
\label{sec:external_knowledge}

\begin{figure}
    \centering
    \includegraphics[width=0.7\linewidth]{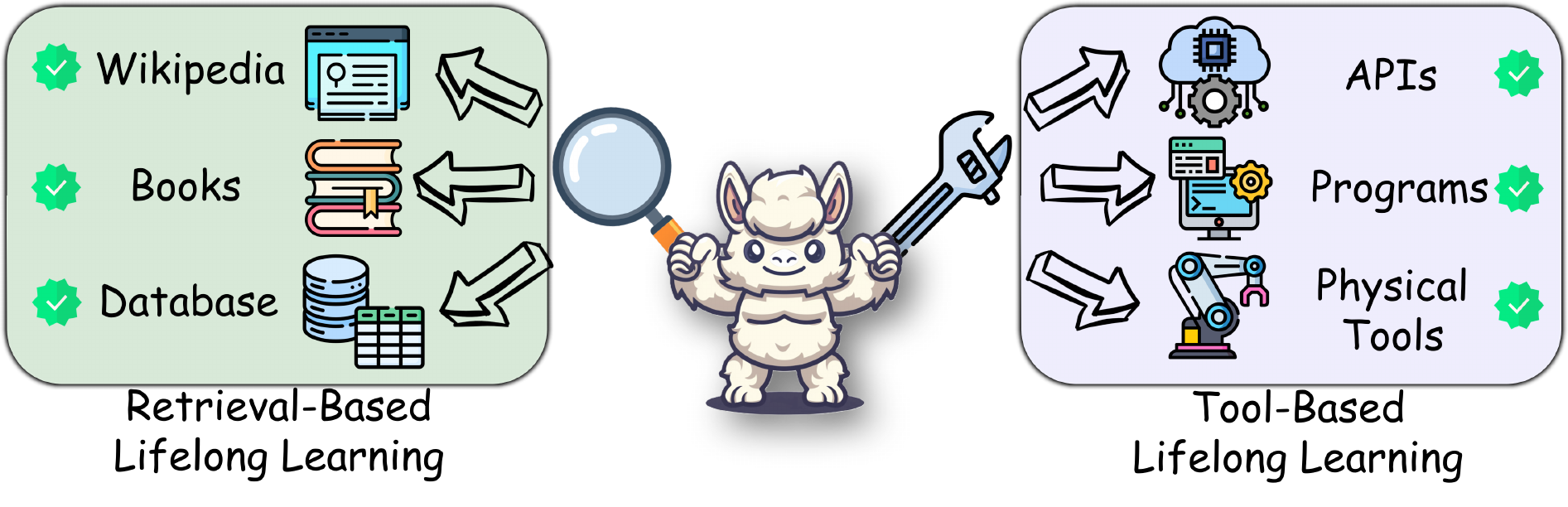}
    \caption{An illustration of two lifelong learning scenarios which equip LLMs with external knowledge: Retrieval-Based Lifelong Learning (left) and Tool-Based Lifelong Learning (right).}
    \label{fig:external_knowledge_illustration}
\end{figure}

Continual pretraining and finetuning are essential for the lifelong learning of LLMs. However, as LLMs grow larger and more powerful, two emerging directions have gained popularity for equipping LLMs with novel external knowledge without modifying their parameters. This survey considers Retrieval-Based and Tool-Based Lifelong Learning, as both are promising approaches for achieving lifelong learning in LLMs. An illustraction is provided in Figure \ref{fig:external_knowledge_illustration}.

\subsection{Retrieval-Based Lifelong learning}
\label{sec:retrieval_based_lifelong_learning}

\noindent \textbf{Why do LLMs need retrieval?} \quad Retrieval-based lifelong learning addresses the critical need for large language models to access and incorporate up-to-date knowledge from external sources \cite{Densepassage,Interleavingretrievalwith,Self-rag}. As the world's information continues to expand and evolve rapidly, static models trained on historical data quickly become outdated, unable to comprehend or generate content about new developments. For example, consider a scenario where a significant medical breakthrough is announced after the model's last training update. In such cases, accessing real-time information from comprehensive databases or continuously updated platforms like Wikipedia becomes invaluable. These external sources offer a vast reservoir of current knowledge, presenting a vital complementary asset to enhance the static nature of pre-trained LLMs \cite{wu2024continual,zhang2023large}.

\noindent \textbf{How to retrieve?} \quad At the heart of implementing this approach is Retrieval-Augmented Generation (RAG), which synergistically combines the deep learning capabilities of LLMs with the dynamic retrieval of external data. RAG models operate by first fetching relevant information using a retriever component before generating text, thus ensuring the content is both updated and contextually appropriate. This process not only enriches the model's output but also significantly extends its applicability to newer domains and topics.
We introduce several approaches that underscore the adaptability and effectiveness of retrieval-based methods as follows: 
Dense Passage Retrieval (DPR) \cite{Densepassage} optimizes the retrieval process by encoding both queries and documents in a dense vector space, allowing for more accurate semantic matching.
Interleaved Retrieval guided by Chain-of-Thought (IRCOT), as proposed by Trivedi et al. \cite{Interleavingretrievalwith}, embeds the retrieval step within the generative process. This approach dynamically adjusts the information retrieved as the response is being formed, which is particularly beneficial in complex dialogues or multi-turn interactions.
Tree of Clarifications (TOC) developed by Kim et al. \cite{Treeofclarifications} structures retrieved knowledge in a hierarchical tree format, enabling precise and relevant information retrieval at varying levels of query complexity.
Active Retrieval in the form of Forward-Looking Active Retrieval augmented generation (FLARE) by Jiang et al. \cite{Activeretrievalaugmented} proactively updates the retrieval database to include the latest information, ensuring the model's responses are timely and informed.
Self-Reflective Retrieval-Augmented Generation (Self-RAG) by Asai et al. \cite{Self-rag} utilizes a feedback loop where the model's output directly influences and refines future retrieval queries, promoting continuous self-improvement.

\subsection{Tool-Based Lifelong Learning}
\label{sec:tool_based_lifelong_learning}

\noindent \textbf{Why do LLMs need tools?} \quad Tool-based lifelong learning for large language models (LLMs) stems from the necessity to extend their functionality beyond static knowledge and enable them to interact dynamically with their environment \cite{qin2023tool,qin2023toolllm,huang2024towards}. In real-world applications, it is often crucial for models to perform tasks that involve operations outside of straightforward text generation or interpretation. For example, an LLM tasked with providing real-time financial advice may need to access and process the latest stock market data, use analytical tools to predict trends or interact with databases to fetch client-specific information. Such scenarios require the model not only to understand and generate language but also to utilize external computational tools effectively, mirroring human capability in using tools to enhance cognitive tasks \cite{allen2020rapid}.

\noindent \textbf{How to use tools?} \quad The development of tool-equipped LLMs, often referred to as ``tool learning'', transforms these models from static repositories of knowledge to dynamic systems capable of performing complex computational tasks and interacting with various APIs and software environments. This transformation is made possible through frameworks designed to teach LLMs how to integrate and utilize different tools effectively. For instance, Chameleon \cite{lu2024chameleon} synthesizes programs to tackle complex reasoning tasks by leveraging a combination of LLMs, visual models, search engines, and custom Python functions. Similarly, the ToolAlpaca framework \cite{tang2023toolalpaca} generates a diverse tool-use corpus through a multi-agent simulation environment, enhancing the model's general tool-use capabilities. Other notable frameworks include Confucius \cite{gao2024confucius}, which employs a multi-stage learning process coupled with feedback mechanisms to refine the tool-using proficiency of LLMs and GPT4Tools \cite{yang2024gpt4tools}, which integrates multiple external tools to expand the functional reach of pre-trained models. Additionally, more complex tool datasets like APIBench \cite{patil2023gorilla} and ToolBench \cite{qin2023toolllm} have been developed to provide a structured environment for training and evaluating the tool-using capabilities of LLMs, broadening the boundary of what these models can achieve in practical applications.

\subsection{Summary}
Building on continual pretraining and finetuning, which enhance LLMs' internal knowledge, equipping LLMs with external knowledge through retrieval-based and tool-based lifelong learning significantly extends their capabilities. Retrieval-based methods ensure models remain updated by incorporating real-time information. Tool-based approaches enable LLMs to interact with external computational tools and APIs. Despite advancements, challenges persist in integrating these techniques seamlessly and efficiently. Future research should focus on refining retrieval mechanisms, improving tool integration frameworks, and developing comprehensive benchmarks to evaluate the effectiveness of external knowledge incorporation in LLMs.

\section{Discussion and Conclusion} \label{sec:discussion_conclusion} 

\subsection{Existing Challenges}
The journey towards optimizing lifelong learning for large language models faces a number of significant challenges that stem from the fundamental characteristics of these systems:
\begin{itemize}
    \item \textbf{Catastrophic Forgetting}: This is a core challenge in lifelong learning, where newer information can overwrite what the model previously learned. As LLMs are continuously updated with new data, ensuring that they retain valuable knowledge from past training without losing it to new and possibly unrelated information remains a critical issue \cite{mccloskey1989catastrophic}.
    \item \textbf{Plasticity-Stability Dilemma}: Finding the right equilibrium between plasticity (the ability to learn new information) and stability (the ability to retain old information) is crucial \cite{mermillod2013stability}. This balance impacts the model’s capacity to acquire domain-specific knowledge, such as medical information while preserving its broad-based general abilities. Additionally, the concept of alignment tax \cite{lin2023mitigating} highlights the challenge in training LLMs to align with human values without compromising their capabilities in areas like reasoning and planning. The objective is to enhance safety and alignment with ethical norms without diluting the model’s functional effectiveness.
    \item \textbf{Expensive Computation Cost}: The computational demand of fully finetuning LLMs, especially for models with billions of parameters, can be prohibitively high. 
    \item \textbf{Unavailability of Model Weights or Pretraining Data}: Often, the original training data or model weights are not available for further refinement due to privacy concerns \cite{lermen2023lora,zhan2023removing}, proprietary restrictions, or commercial licenses. 
\end{itemize}

\subsection{Current Trends}

As highlighted by the existing challenges, the evolution of lifelong learning for large language models is significantly influenced by the high computational costs of training these models and their robust capabilities. This has led to several new trends in how lifelong learning is approached: 
\begin{itemize}
    \item \textbf{From Specific to General Tasks}: There is a noticeable shift from focusing on specific tasks like text classification \cite{ke2021classic} and named entity recognition \cite{monaikul2021continual} to more general tasks that expand the model's utility across different domains. This transition towards general tasks such as instruction tuning \cite{chen2024coin} and knowledge editing \cite{CKEsurvey} leverages the broad generalization ability of LLMs, allowing them to handle diverse challenges without intensive retraining for each specialized task.
    \item \textbf{From Full to Partial Finetuning}: Considering the substantial resources required to fully finetune LLMs, there is a growing preference for partial finetuning strategies. Techniques like Adapter layers \cite{houlsby2019parameter}, Prompt tuning \cite{lester2021power}, and LoRA \cite{hu2021lora} adjust only a small subset of parameters, preserving the core model while integrating the flexibility to adapt to new data and tasks efficiently.
    \item \textbf{From Internal to External Knowledge}: To overcome the limitations of frequent internal updates, there is a notable trend towards employing external knowledge sources. Strategies such as Retrieval-Augmented Generation \cite{knowledge-intensivenlp} and tool-based learning \cite{qin2023tool} enable LLMs to access and utilize current external data dynamically. This approach not only enhances the model’s problem-solving capacity but also ensures continual learning with minimal retraining.
\end{itemize}

\subsection{Future Directions}

As the capabilities of LLMs become stronger, the computational costs increase, and their applications broaden, future lifelong learning will aim to equip LLMs with more general abilities beyond the text modality, reduce computational costs, and address more realistic scenarios. Here are three promising areas of focus that could significantly advance the field: 
\begin{itemize}
    \item \textbf{Multimodal Lifelong Learning}: The integration of multiple modalities beyond text—such as images, videos, audio, time-series data, and knowledge graphs—into lifelong learning paradigms is a burgeoning area of research \cite{he2023continual,chen2024coin}. This approach aims to develop more comprehensive and versatile models that can process and understand a broader array of data types, mirroring human-like learning capabilities across various sensory inputs.
    \item \textbf{Efficient Lifelong Learning}: To manage the computational demands of training and updating LLMs, researchers are looking towards more efficient strategies. These include leveraging model pruning \cite{sun2023simple} to eliminate unnecessary parameters, model merging \cite{goddard2024arcee} to consolidate knowledge, and model expansion \cite{llama,solar} to adaptively increase capacity without extensive retraining. Additionally, capitalizing on the in-context learning abilities of state-of-the-art LLMs, which support extensive contexts up to 10 million tokens, is seen as highly promising. For example, Gemini 1.5 Pro \cite{reid2024gemini} showcases the potential by translating languages with high accuracy with only reference materials, mimicking a human learning context.
    \item \textbf{General Lifelong Learning}: The ultimate goal in this field is to enable LLMs to actively acquire new knowledge and learn through dynamic interactions with their environments, rather than solely from static datasets \cite{wang2024survey}. Incorporating principles from reinforcement learning, agent-based systems, and embodied AI could lead to the development of truly general AI. This ambitious direction seeks to emulate the natural lifelong learning capabilities of humans, facilitating a deeper, more intuitive engagement with the world.
\end{itemize}

\subsection{Conclusion}
In conclusion, this survey systematically categorizes existing studies into 12 lifelong learning scenarios and provides a comprehensive exploration of the methodologies. Our analysis highlights the delicate balance required to manage catastrophic forgetting, ensure computational efficiency, and maintain a balance between specificity and generality in knowledge acquisition. As the field continues to evolve, the integration of these advanced strategies will undoubtedly play a crucial role in shaping the next generation of AI systems, helping them closer to achieving true, human-like learning and adaptability.

\begin{acks}
The work described in this paper was partially funded by the National Natural Science Foundation of China (Grant No. 62272173), the Natural Science Foundation of Guangdong Province (Grant Nos. 2024A1515010089, 2022A1515010179), and the Science and Technology Planning Project of Guangdong Province (Grant No. 2023A0505050106). The icons used in this paper are downloaded from https://www.flaticon.com/ and are created by Iconjam, Freepik, Whitevector, Eucalyp, and Pixel perfect.
\end{acks}

\bibliographystyle{ACM-Reference-Format}
\bibliography{main}

\end{document}